\pdfoutput=1
\documentclass[11pt]{article}

\usepackage[final]{acl}

\usepackage{times}
\usepackage{latexsym}

\usepackage[T1]{fontenc}
\usepackage{multicol,lipsum}
\usepackage[utf8]{inputenc}

\usepackage{microtype}

\usepackage{inconsolata}

\usepackage{graphicx}

\usepackage{CJKutf8}

\usepackage{float} % 需要这个包来使用 [H] 选项
\usepackage{amsmath}
\usepackage{multirow}
\usepackage{multicol} % 处理多栏
\usepackage{caption} % 处理标题
\usepackage{url}
\usepackage{subcaption}
\usepackage{booktabs} % 用于生成漂亮的表格
\usepackage{array} % 用于控制表格列宽

\usepackage{cleveref}
\title{Psy-Insight: Explainable Multi-turn Bilingual Dataset for Mental Health Counseling }
\author{First Author \\
  Keqi Chen / ckqqqq@bupt.edu.cn \\
  \texttt{email@domain} \\\And
  Second Author \\
  Zekai Sun / sunzekai@bupt.edu.cn \\
  Yuhua Wen / sunzekai@bupt.edu.cn \\
  Affiliation / Address line 1 \\
  Affiliation / Address line 2 \\
  Affiliation / Address line 3 \\
  \texttt{email@domain} \\}

\author{
  Keqi Chen\thanks{Equal Contribution}, Zekai Sun\footnotemark[1], Yuhua Wen, Huijun Lian, Yingming Gao, Ya Li\\
  Beijing University of Posts and Telecommunications, Beijing \\
  \texttt{\{ckqqqq,sunzekai,yuhuawen,lhj,liqifei,yingming.gao,yli01\}@bupt.edu.cn}
}
\begin{document}
\maketitle
% \begin{abstract}
% This document is a supplement to the general instructions for *ACL authors. It contains instructions for using the \LaTeX{} style file for 2 2023. 
% The document itself conforms to its own specifications, and is, therefore, an example of what your manuscript should look like.
% These instructions should be used both for papers submitted for review and for final versions of accepted papers.
% \end{abstract}

% \section{Introduction}

% These instructions are for authors submitting papers to EMNLP 2023 using \LaTeX. They are not self-contained. All authors must follow the general instructions for *ACL proceedings,\footnote{\url{http://acl-org.github.io/ACLPUB/formatting.html}} as well as guidelines set forth in the EMNLP 2023 call for papers. This document contains additional instructions for the \LaTeX{} style files.
% The templates include the \LaTeX{} source of this document (\texttt{EMNLP2023.tex}),
% the \LaTeX{} style file used to format it (\texttt{EMNLP2023.sty}),
% an ACL bibliography style (\texttt{acl\_natbib.bst}),
% an example bibliography (\texttt{custom.bib}),
% and the bibliography for the ACL Anthology (\texttt{anthology.bib}).

\begin{abstract}

The in-context learning capabilities of large language models (LLMs) show great potential in mental health support. However, the lack of counseling datasets, particularly in Chinese corpora, restricts their application in this field. To address this, we constructed Psy-Insight, the first mental health-oriented explainable multi-task bilingual dataset. We collected face-to-face multi-turn counseling dialogues, which are annotated with multi-task labels and conversation process explanations. Our annotations include psychotherapy, emotion, strategy, and topic labels, as well as turn-level reasoning and session-level guidance. Psy-Insight is not only suitable for tasks such as label recognition but also meets the need for training LLMs to act as empathetic counselors through logical reasoning. Experiments show that training LLMs on Psy-Insight enables the models to not only mimic the conversation style but also understand the underlying strategies and reasoning of counseling. Our code, expert evaluation results, and the Psy-Insight dataset have been open-sourced \footnote{\url{https://ckqqqq.github.io/Demo/Psy-Insight/}}. 

\end{abstract}

\section{Introduction}

According to a report from the World Health Organization \cite{who2022mental}, 71\% of individuals in low-income countries suffering from mental disorders are unable to receive timely treatment. Patients with mental disorders require affordable and easily accessible mental health support. Mental support chatbots offer a possible solution\cite{liu2023chatcounselor}.

% Previous research has made significant efforts in developing Chatbots specifically designed for mental health support, bringing hope for cheap and reachable AI-assisted mental health support. However, previous psychological dialogue systems, most relying on rigid pipeline modules like Named Entity Recognition(NER), Knowledge Graphs(KG), and emotion classification, sometimes failed to adapt to diverse real-life psychological counseling scenarios.

General-purpose large language models (LLMs) such as ChatGPT \cite{chatgpt_website} and Bard \cite{bard_website} have shown their potential in various conversations, which include enhanced in-context learning\cite{ICT-incontext-learning2022} and chain-of-thought\cite{wei2022COT} capabilities for multiple tasks. This allows LLMs to effectively handle the complex situations encountered in real-life multi-turn counseling.

For mental-supported LLMs, there is a need for suitable dialogue datasets for finetuning. However, the potential of large models has not been fully explored on previous counseling datasets.~\cite{tradiDatatset,tradidataset2}. These datasets are characterized by short labels for single-task annotations, such as emotion or entity labels, which are designed for subtasks within the traditional pipeline chatbots, such as emotion classification~\cite{emotionclassificaition2023} 
named entity recognition~\cite{NER2007}, and knowledge graph completion~\cite{peng2022KG_dialog}.

However, the strength of LLMs lies not only in their ability to recognize and classify but also in their potential to reason step-by-step. They can capture analysis and explanation from therapists beyond what is described by short labels (eg., “Beth fears that sharing could worsen things” vs “Fear”).  Therefore, when designing datasets for mental support LLMs, these capabilities should be reflected in the datasets. The focus should not only be on short labels for simple tasks but also on reasoning annotations that describe the thought process for complex task scenarios.

\begin{figure*}[ht]
    \centering
    \includegraphics[width=0.92\linewidth]{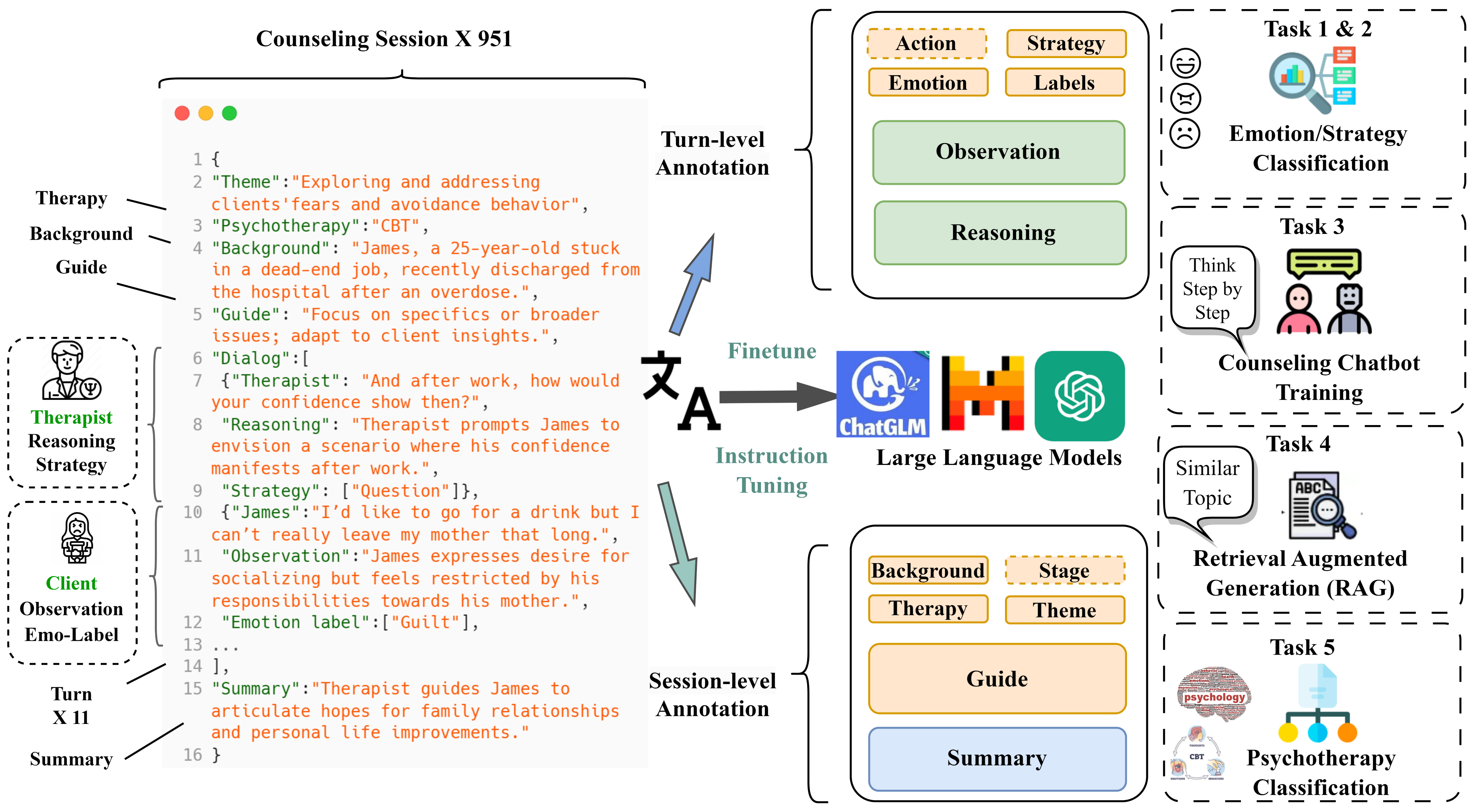}
    \vspace{-0.3cm}
    \caption{The left section presents Psy-Insight's counseling dialogues and annotations, while the right section illustrates the corresponding multi-tasks for these annotations. The Psy-Insight dataset features 951 sessions of multi-turn counseling dialogues annotated with step-by-step reasoning and multi-task labels. Within a session, the therapist and client engage in 5\~60 turns of dialogue on a single topic. We have annotated counseling dialogues at various granularity levels. The example of Chinese data is shown in Table~\ref{tab:chinese_example}}
    \label{fig: PsyInsight_example}
\end{figure*}

Motivated by the goals, we constructed the \textbf{Psy-Insight}, the first \textbf{Bilingual} corpus of \textbf{Explainable Multi-turn Counseling}. Our dataset includes 520 sessions of English multi-turn counseling and 431 sessions in Chinese. These multi-turn dialogues can help LLMs imitate human face-to-face counseling. Moreover, the \textbf{explainable annotations} in Psy-Insight can help LLMs understand the analysis and logic behind counseling.

% \label{section:my}
The examples from the Psy-Insight dataset are presented in Table~\ref{tab:english_example}, and Table~\ref{tab:chinese_example}. Dialogue and annotations in Psy-Insight are structured hierarchically, from broad to specific: Case - Session - Turn.

Previous studies \cite{2021multitask} showed that explicit multitask learning can improve the generation ability of LLMs. So, the Psy-Insight collected labels of counseling across a range of psychological task scenarios. The counseling data in Psy-Insight, collected from blogs and books, is annotated with multitask labels and step-by-step processes at both turn-level and session level. 

As shown in Figure~\ref{fig: PsyInsight_example}, Psy-Insight features 5 concise labels for multi-task learning~\cite[MTL]{2021multitask} and 6 descriptive annotations for LLMs analysis and reasoning. Concise Labels include emotion labels (e.g., Guilt), psychotherapy method labels (e.g., Cognitive Behavioral Therapy), strategies (e.g., Questions), and topics (e.g., Academic Pressure). Additionally, Psy-insight also includes descriptive annotations for analysis and reasoning, such as background, guide, summary, theme, and so on. As shown in Figure~\ref{fig: PsyInsight_example}, these annotations make the Psy-Insight dataset suitable for at least five psychological tasks in NLP, such as emotion/psychotherapy/strategy classification, retrieved argument generation \cite[RAG]{2020RAG}, and dialogue generation.

% \vspace{-0.0cm}
 We have conducted finetuning and RAG experiments on Psy-Insight. The automatic and expert evaluations have shown that Psy-Insight can enhance the performance of mental support LLMs. Our contributions are summarized as follows:
\begin{itemize}
    \item We construct a bilingual, explainable multi-turn counseling corpus, Psy-Insight, for easily training mental support LLMs. 
    \item Psy-Insight includes step-by-step reasoning and multi-task labels at both session and turn levels. These annotations meet LLMs' requirements for chain-of-thought and multi-task learning. We trained LLMs on the Psy-Insight dataset, and our results showed that it enhances their performance in mental support.
    \item  We invited mental health experts to compare our dataset with baseline datasets. Expert reviews show that our dataset is of high quality. The results of the expert evaluation can also be used for further research.
\end{itemize}

% \begin{figure*}[t]
%   \includegraphics[width=0.48\linewidth]{example-image-a} \hfill
%   \includegraphics[width=0.48\linewidth]{example-image-b}
%   \caption {A minimal working example to demonstrate how to place
%     two images side-by-side.}
% \end{figure*}

% \begin{figure}[t]
%   \includegraphics[width=\columnwidth]{example-image-golden}
%   \caption{A figure with a caption that runs for more than one line.
%     Example image is usually available through the \texttt{mwe} package
%     without even mentioning it in the preamble.}
%   \label{fig:experiments}
% \end{figure}

% \begin{figure*}[h]
%   % \includegraphics[bb=0 0 400 300,width=\columnwidth]{Dialog.png}
%   \includegraphics[width=1\linewidth]{Picture/Dialog.png}
%   % \DeclareGraphicsExtensions{.jpg}
%   \caption {A minimal working example to demonstrate how to place
%     two images side-by-side.}
% \end{figure*}

\begin{figure*}[ht]

  \includegraphics[width=1\linewidth]{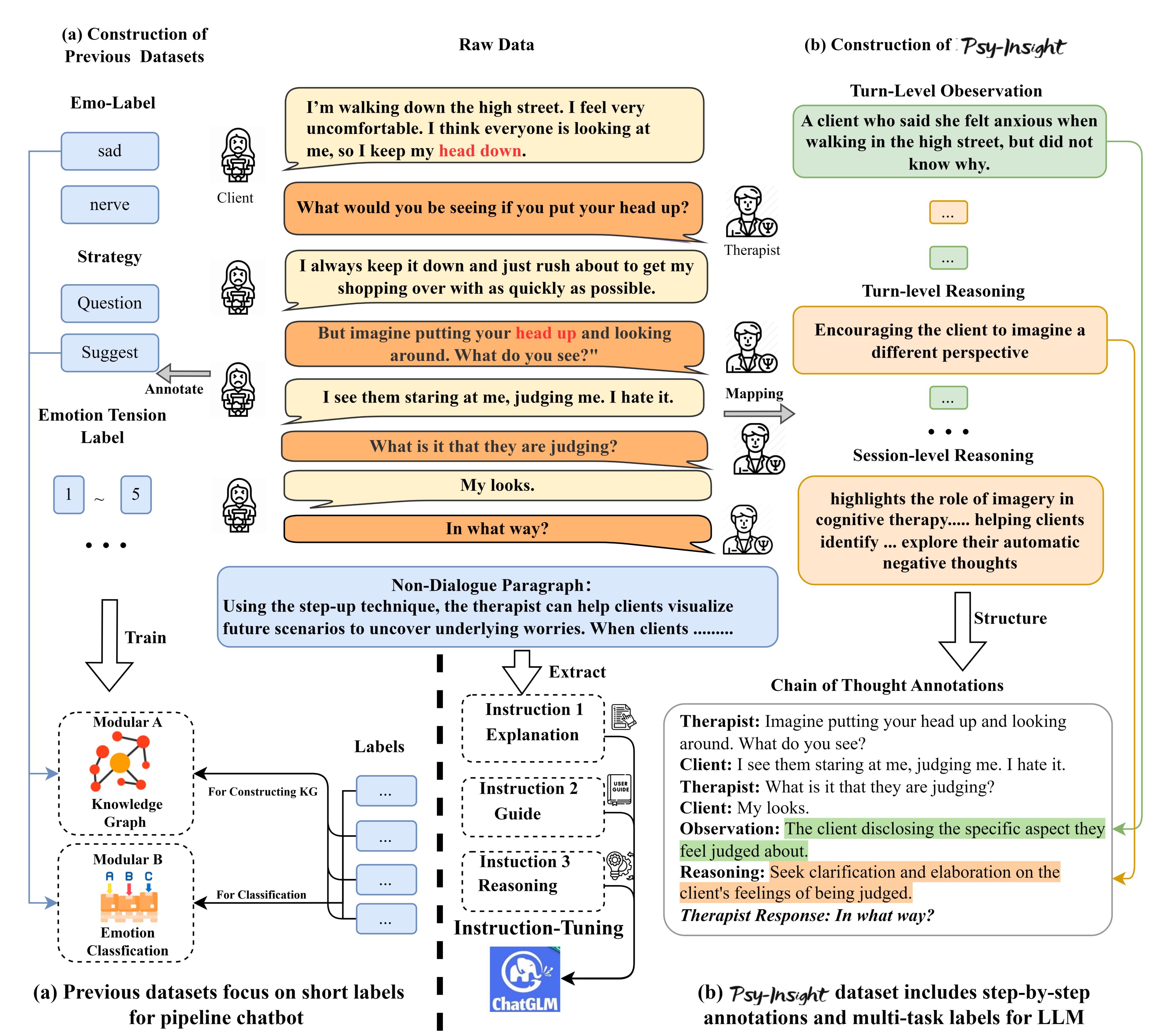}
  % \DeclareGraphicsExtensions{.jpg}
  % \label{fig:comp}
  \vspace{-0.7cm}
  \caption {Comparison of the construction processes of previous datasets and Psy-Insight dataset. As shown on the left side, previous datasets primarily focused on annotating short labels, which are suitable for the subtask in pipeline. The Psy-Insight dataset shown on the right side emphasizes the interpretability of the dialogue process, with step-by-step reasoning and session-level guide and explanation. We also collect multi-tasks labels based on each dialog sessions, offering data to enhance LLM's generalization ability. }
  \label{fig: comp}
\end{figure*}
\section{Related Work}

We focus on multi-turn dialogue data for mental support, which is related to research on emotional chatbots and psychological datasets.

\subsection{Mental Support Chatbots}

Some researchers~\cite{vaidyam2019chatbots,liu2018chatbot,kretzschmar2019chatbot,smith2018chatbot,huang2018ECM} first evaluated the ability of chatbots to offer mental support.~\citet{majumder-etal-2020-mime} tried to offer emotional support by mimicking clients' emotions. Advances in language models have led researchers such as \citet{cheng2022—multiESC,peng2023fadoESC,tu2022misc,peng2022KG_dialog} to integrate strategic components like knowledge graphs, emotion classifiers, and strategy encoders to guide chatbots in mental support response through pipeline modular. 

Following ChatGPT's emergence, the capabilities of LLMs have attracted researchers to apply them to psychological tasks. Studies by \citet{elyoseph2023chatgpt,li2023chatgpt} show that LLMs excel in emotional chat, with finetuning in high-quality datasets. Researchers \cite{chen2023soulchat,qiu2023smile} finetuned LLMs on emotional datasets. Their private LLMs outperformed traditional pipeline chatbots in mental support.

\subsection{Dialogue Dataset for Mental Support }

Early research \cite{chadha2021suicide,yao2022depression} attempted to crawl emotional dialogue from sources like tweets for the classification of depressive or suicidal tendencies. Some researchers endeavored to collect counseling data, such as the Empathetic Dialogue \cite{rashkin2018Emphatheicdialog} and Esconv \cite{liu2021Esconv}. 

There is a shortage of psychological counseling dialogue data in Chinese, especially high-quality multi-turn counseling conversations. Influential and high-quality dataset PsyQA~\cite{sun2021psyqa} includes various cases in real life but is limited to single-turn style. It is unsuitable for simulating face-to-face psychological counseling. Researchers introduced two synthetic datasets, SMILE~\cite{qiu2023smile} and SoulChat~\cite{chen-etal-2023-soulchat}, which utilize ChatGPT to expand single-turn dialogues in PsyQA into multi-turn conversations.

% \subsection{Transformation}

% \begin{table}[htbp]
% \centering
% \caption{Comparison Between Our Dataset and Related dataset}
% \label{tab:dataset_summary}
%     \begin{tabular}{p{1.5cm}p{1.7cm}p{1.2cm}cp{1cm}p{1.3cm}p{1.5cm}c}
%     \toprule
%     Dataset & Domain & Data Type & Avg. Turn & Total & Language & Annotation & Public\\
%     \midrule
%     MotiVAte & Mental Health  & authentic & 1     & 25  & English & - & yes\\
%     ESConv & Emotional Support  & authentic & 20.7& 18 & English & - & yes\\
%     MedDialog & Medical Dialogue& authentic & 12 & 18 & English & - & yes\\
%     D^{4} & Depression Diagnosis& authentic & 21.55 & 18 & Chinese & - & yes\\
%     Misc & Emotional Dialogue& authentic & 21.55 & 7293 & Chinese & - & no\\
%     CTT & Mental Health& authentic & 21.55 & 200 & English & - & no\\
%     PsyQA \cite{sun2021psyqa}  & Mental Health  & authentic & 1 & 20000  & Chinese & Emotional Label& yes\\
%     Smile & Mental Health  & authentic & 23 & 200000                       & Chinese & -& yes\\
%     SoulChat & Mental Health  & synthetic & 38 & 200000                    & Chinese & -& yes\\
%     \textbf{Psy-Insight} & Mental Health & synthetic & 46 & 5334(Cn)\ 5231(En)  & Chinese English & Explanation& yes\\
%     \bottomrule
%     \end{tabular}
% \end{table}

% Using the step-up technique, the therapist can help clients visualize future scenarios to uncover underlying worries

As shown in Figure ~\ref{fig: comp}, the mental support chatbots are shifting from pipeline-based models to LLMs. This implies the need for more multi-turn dialogue datasets designed specifically for LLMs. Synthetic dataset fills gaps in the volume of multi-turn counseling data for pre-training, but they still have significant differences compared to human counseling in quality. 

As detailed in Appendix~\ref{sec:comparison}, we invited psychological experts and students to compare and score ChatGPT-synthetic and human counseling sessions (Table \ref{tab:Psy-insight-vs-smile}). The experts' comments indicate that AI-synthesized counseling texts exhibit shallow empathy, less emotional interaction, and insufficient emotional attention. In contrast to human therapists, LLMs tend to overemphasize problem-solving and fail to fully engage the seeker's intrinsic help-seeking motivation. This reveals the weaknesses of LLMs in establishing deep emotional connections and exploring underlying motivations.

\section{Psy-Insight Construction}

The goal of the Psy-Insight dataset is to collect \textbf{non-synthetic multi-turn bilingual counseling} for training LLMs and synthesize \textbf{step-by-step explanations} for multi-tasks with the help of original explanatory texts. We use synthetic labels to improve the interpretability of datasets while maintaining the quality of dialogues based on real counseling data. Our method aims to fully unlock the potential of real counseling and narrative context in raw data, maximizing their utility.
\begin{figure}[ht]
    \centering
    \includegraphics[width=0.48\textwidth]{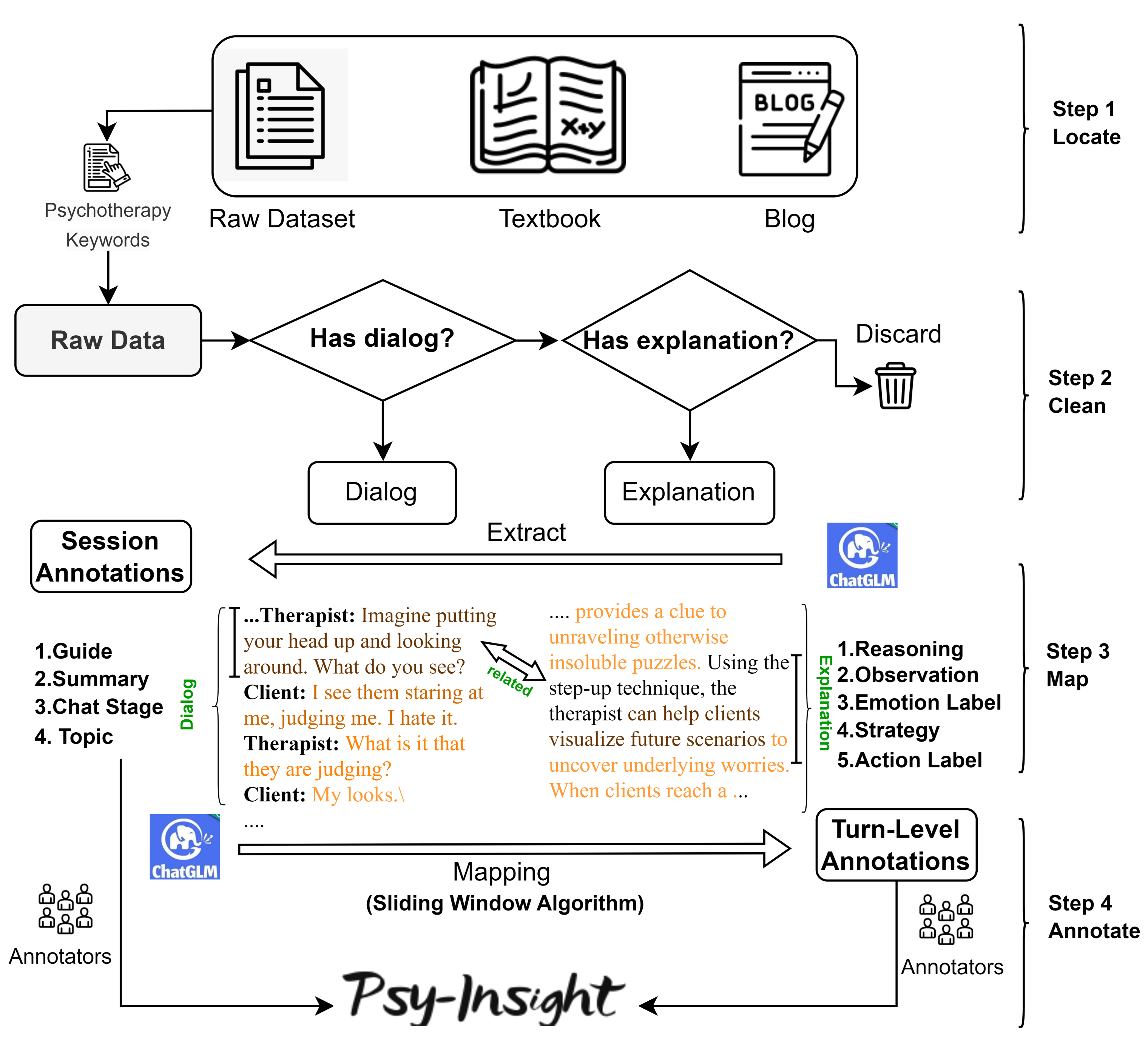}
    \centering
    \caption{The construction workflow of Psy-Insight Dataset. Our workflow involves 4 steps: (1) Locating dialogues and explanation with psychotherapy keywords; (2) Data cleaning; (3) Mapping dialogues and explanation with sliding window algorithm, and computing similarity with embedding models and LLMs; (4) Checking annotations by human annotators. }
    \label{fig:data-workflow}
\end{figure}

\subsection{Workflow}

We collect dialogues from blogs, books, and common crawled websites. The data source records and copyright information are shown in Appendix~\ref{sec:copyright}. Figure~\ref{fig:data-workflow} shows the whole workflow of corpus construction. 

\subsection{Psychotherapy}
Table~\ref{tab:therapy} shows the psychotherapy labels in Psy-Insight. Following suggestions for professional therapists, we utilize these psychotherapy's keywords as anchors (eg. Keyword: Solution-focused Brief Therapy) for crawling and locating counseling cases in raw data. Psychotherapies such as SFBT and CBT focus on case studies, with their dialogue data frequently found in blogs and books. This makes them take a significant part of the whole dataset. These therapy labels are beneficial for psychological research.
% % In Psy-Insight, most of the counseling cases are from modern therapies rather than traditional therapies. Additionally, we also collect some emerging modern psychotherapies such as family therapy.

 \begin{table}[ht]
    \centering

  \resizebox{0.5\textwidth}{!}{ 
    \begin{tabular}{p{4.7cm}rr}
        \toprule
        \textbf{Psychotherapy} & \textbf{Session} & \textbf{Ratio} \\
        \midrule
        CBT\citeyearpar{CBT1987}  & 191 & 24.29\% \\
        REBT\citeyearpar{REBT1996} & 13 & 1.37\% \\
        SFBT\citeyearpar{SFBT1997}  & 398 & 41.85\% \\
        Adlerian Counseling\citeyearpar{Adlerian1953} & 20 & 2.1\% \\
        Client-centered Therapy\citeyearpar{ClientCenter1946} & 28 & 2.94\% \\
        Family Therapy\citeyearpar{FamilyTherapy1984} & 43 & 4.52\% \\
        Gestalt Therapy\citeyearpar{gestalt1993} & 12 & 1.26\% \\
        Multicultural Therapy\citeyearpar{multicultural1991} & 14 & 1.47\% \\
        Postmodern Therapy\citeyearpar{Postmodern1996} & 16 & 1.68\% \\
        Psychoanalytic Therapy\citeyearpar{psychoanalytic1980} & 46 & 4.84\% \\
        Psychodynamic Model\citeyearpar{psychodynamic1994} & 23 & 2.42\% \\
        Reality Therapy\citeyearpar{2010reality} & 10 & 1.05\% \\
        Unknown & 10 & 13.15\% \\
        \textbf{Overall} & \textbf{786} & \textbf{100\%} \\
        \bottomrule
    \end{tabular}
    }
    \vspace{-0.1cm}
    \caption{The distribution for psychotherapy labels in Psy-Insight. Most of the counseling cases are from case-centered modern psychotherapies, such as Solution-Focused Brief Therapy, SFBT\citeyearpar{SFBT1997} and Cognitive Behavioral Therapy, CBT\citeyearpar{CBT1987}, Rational emotive behavior therapy, REBT\citeyearpar{REBT1996}. We only found a few cases in traditional psychotherapies in crawled data (eg., Psychodynamic Therapy \citeyearpar{psydy}).}
    \label{tab:therapy}
\end{table}

\subsection{Text Segmentation}

% As shown in Figure~\ref{fig:data-construction}, dialogues in Psy-Insight are collected from real-life counseling found in textbooks and blogs.
We utilized psychotherapy labels and regex methods to locate the conversation structure and then crawled the text fragment from raw data. In these text fragments, we found numerous descriptive paragraphs that are related to dialogues. These non-dialogue sections contain rich psycho-explanations, including background, therapist reflections, explanations so on. We retained these text for label mapping. 

% We use regulation tools to extract information from the speakers. To protect the privacy of the client and therapist. We transform all specific name into "therapist", "client", "Smith" and "Mary"之类的代称或者常见名称.同时我们删除了对话中提到的隐私数据。在家庭疗法等常见中，对话参与人数会大于两人，同时对话的主体会发生转变，我们对这部分数据进行了特殊的人工处理

 % \textbf{\FirstWord{Other Tags}}

\subsection{Labels Mapping in Sliding Window}

We map descriptive paragraphs into session-level labels and turn-level explanations for related dialogue. Specifically, we use a sliding window algorithm to trace related descriptive paragraphs and dialogue in the order they appear. By adjusting the window size, we can obtain dialog-description key-value pairs at different granularity levels. For example, a descriptive paragraph may explain multiple turns of dialogue within a session. By setting the sliding window, we map the content of a specific dialogue turn with a descriptive text. We then use sentence-transformer\cite{2019-sentence-bert} and GLM4-9B~\cite{du2022glm} to determine if the dialogue turn and its descriptive text are relevant.

These related dialogue-description text pairs include a dialogue window and its corresponding descriptive information. The dialogue window acts as the key, and the descriptive text as the value. If the descriptive text covers multiple dialogue turns, we break it into session-level labels, such as background introductions before first counseling. If the dialogue window size is less than three turns, we break the descriptive text into turn-level labels, such as observations of the client's expressions or emotions. By analyzing the dialogue window length and descriptive content, we convert unstructured paragraphs into labels of varying granularity.

At the end of the workflow, annotators will review and verify the labels for each session. Not all dialogues have related explanatory texts, so some labels are marked as Unknown.

\section{Psy-Insight Analysis}

We collected 189 cases of face-to-face dialogues, with 75 cases in the Chinese dataset. Each case includes several sessions in the counseling cycle, and each session involves several turns of dialogue. 
%  这个应该放到数据描述里面
% Table~\ref{tab:therapy}-~\ref{tab:emolabel} show statistics of the English dataset, Chinese dataset, strategies and emotion labels respectively. 

\begin{figure}[t]
    \centering
    \includegraphics[width=0.47\textwidth]{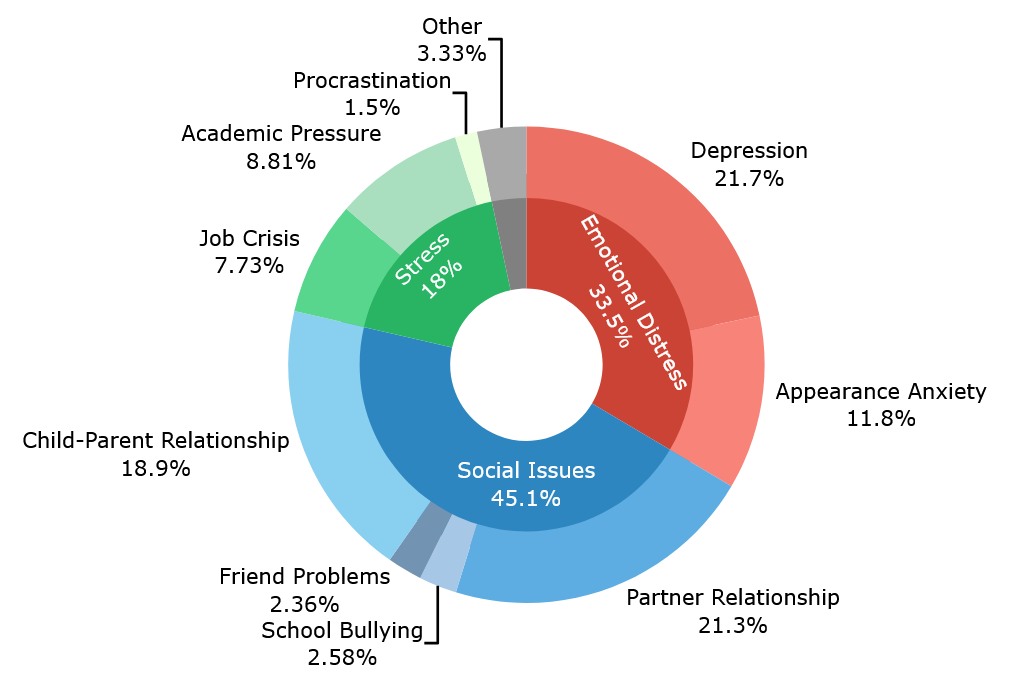}
    \centering
    \caption{Statistics of topics in counseling of Psy-Insight. Top-3 topics are Depression (21.7\%), Partner Relationship (21.3\%), Child-Parent Relationship (18.9\%).}
    \label{fig:topic_distribution}
\end{figure}

\subsection{Conversation Statistics in Psy-Insight}

\begin{figure*}[!ht]
    \centering
    \begin{subfigure}[b]{0.47\linewidth}
    \includegraphics[width=\linewidth]{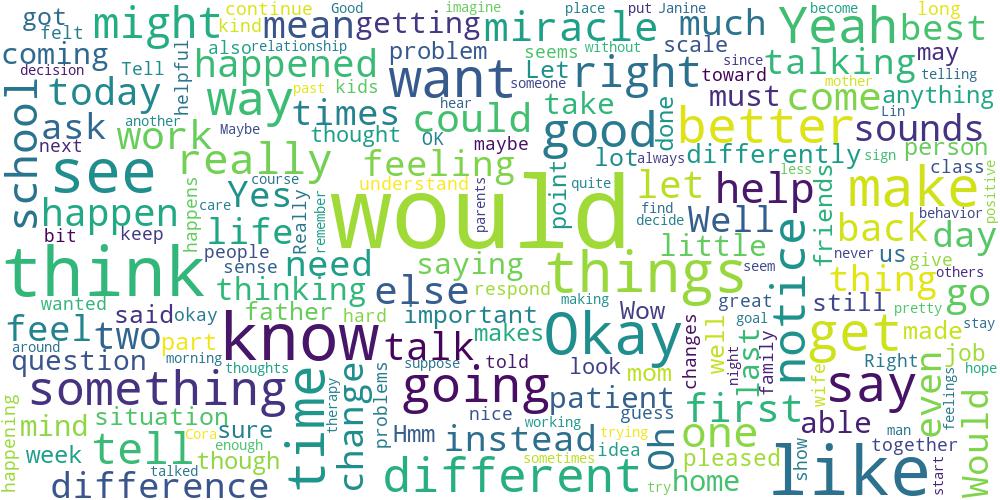}
    \caption{Top-100 common words in English counseling.}
    \label{fig:enTop100}
    \end{subfigure}
\hfill
    \begin{subfigure}[b]{0.47\linewidth}
    \includegraphics[width=\linewidth]{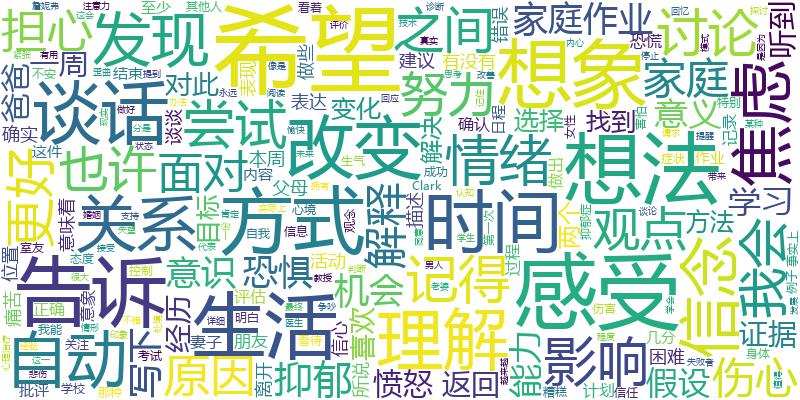}
    \caption{Top-100 frequent words in Chinese counseling}
    \label{fig:cnTop100}
    \end{subfigure}
    \caption{Word cloud figures of Chinese and English counseling.}
    \label{fig:fig3}
\end{figure*}

Table~\ref{tab:EnglishD}-\ref{tab:ChineseD} and Figure~\ref{fig:fig3} shows statistical data of counseling in Psy-Insight.

\noindent\textbf{Face-to-Face Counseling}
Compared to GPT-synthetic~\cite{qiu2023smile,chen2023soulchat} datasets, the conversations in Psy-Insight have longer average dialogue turns (55 vs 24 turn/case). As shown in Table~\ref{tab:compare_with_datasets}, human therapists tend to respond with shorter turns to help and guide their patients in a multi-turn interaction ( 1.7 sentence/response ). However, ChatGPT often gives long suggestions straight away, compared to shorter conversations ( avg. 10 turn/response in the Smile dataset).

% At the same time, LLM is more likely to provide long and useful suggestions to solve problems directly.  

\noindent\textbf{Topics} Figure \ref{fig:topic_distribution} shows the topic distribution in Psy-insight, over 50\% topics belong to top-3 topics 

\noindent\textbf{High-frequency Words} We also analyze the high-frequency word distribution in Psy-Insight 's dialog. Figure~\ref{fig:enTop100} displays the top 100 English consulting words, while Figure~\ref{fig:cnTop100} does the same for Chinese. Appendix \ref{sec:Statistics} details high-frequency words in the annotations.

\noindent\textbf{Bilingual Corpus} Our statistics reveal that high-frequency words and topics show notable similarities in both Chinese and English conversations, with the same words accounting for as much as 42\%. Our bilingual dataset holds potential value for cross-cultural psychological research.

\begin{table}[ht!]
\centering
    \resizebox{0.47\textwidth}{!}{
    \begin{tabular}{lccc}
    \toprule
    \textbf{Category} & \textbf{Total} & \textbf{Therapist} & \textbf{Client} \\
    \hline
    Cases & 114 & - & - \\
    Sessions & 520 & - & - \\
    Turn & 6,208 & 3,147 & 3,151 \\
    Avg. Utts. / Case & 54.46 & 27.15 & 27.31 \\
    Avg. Utts. / Session  & 11.94 & 5.95 & 5.99 \\
    Reasoning &3,061 & 3,061 & - \\
    Observation & 3,090& -& 3,090 \\
    \bottomrule
    \end{tabular}
    }
\caption{Statistics of English Counseling in Psy-Insight. Dialogue in Psy-Insight is organized into different levels, ranging from large to small: Case - Session - Turn. During the whole counseling cycle in a specific case, the therapist always engages in a session for a topic. Each session consists of dozens of dialogue turns.}
\label{tab:EnglishD}
\end{table}

\begin{table}[ht!]

\centering
    \resizebox{0.47\textwidth}{!}{
    \begin{tabular}{lccc}
    \toprule
    \textbf{Category} & \textbf{Total} & \textbf{Therapist} & \textbf{Client} \\
    \hline
    Cases & 75 & - & - \\
    Sessions & 431 & - & - \\
    Turn & 5,776 & 2,911 & 2,895 \\
    Avg. Utts. / Case & 77.01 & 38.81 & 38.2 \\
    Avg. Utts. / Session & 13.4 & 6.75 & 6.65 \\
    Reasoning & 2,831 & 2,831 & - \\
    Observation & 2,797 & -& 2,797 \\
    \bottomrule
    \end{tabular}
    }
\caption{Statistics of Chinese Counseling in Psy-Insight. }

\label{tab:ChineseD}
\end{table}

\begin{table}[ht]
    \centering
    \resizebox{0.47\textwidth}{!}{
        \begin{tabular}{p{4cm}rr}
            \toprule
            \textbf{Therapist's Strategy} & Num & Proportion \\
            \midrule
            % \multicolumn{3}{l}{\textbf{Therapist's Strategy}} \\
            Reassurance & 168 & 2.77\% \\
            Information & 139 & 2.29\% \\
            Unknown & 551 & 10.73\% \\
            Providing Suggestions & 122 & 2.01\% \\
            Question & 3,896 & 64.14\% \\
            Reflection of Feelings & 868 & 14.29\% \\
            Restatement & 121 & 1.99\% \\
            Role-play & 89 & 1.47\% \\
            Self-disclosure & 19 & 0.31\% \\
            \textbf{Overall} & \textbf{5,973} & \textbf{100\%} \\
            \midrule
            
            {\textbf{Client's Emotion}} & Num & Proportion \\
            \midrule
            % \multicolumn{3}{l}{\textbf{Client's Emotion}} \\
            Anger & 622 & 9.03\% \\
            Anxiety & 912 & 13.23\% \\
            Depression & 425 & 6.17\% \\
            Fear & 340 & 4.93\% \\
            Guilty & 248 & 3.6\% \\
            Happiness & 867 & 12.58\% \\
            Neutral & 1,777 & 25.79\% \\
            Unknown & 319 & 15.85\% \\
            Sadness & 493 & 7.15\% \\
            Shame & 115 & 1.67\% \\
            \textbf{Overall} & \textbf{6,118} & \textbf{100\%} \\
            \bottomrule
        \end{tabular}
    }
    \caption{Label distribution for emotional labels and therapist's strategies in Psy-Insight dataset.}
    \label{tab:emolabel}
\end{table}

% RESULT REUSLT

% RESULT REUSLT
%  ATTEIONTION !!! 

% CBT Cognitive Behavior Therapy
% SFBT Solution-Focused Brief Therapy
% Rational Emotive Behavior Therapy REBT

\subsection{Label Analysis in Psy-Insight}
As shown in Table ~\ref{tab:therapy},\ref{tab:emolabel}, we also analyze the multi-task labels and other explainable labels. Concise labels are suitable for multi-task learning (eg. Classification). Session-level descriptive annotations such as topics and guidance are suitable as cases' keys for LLM generation (eg. retrieved argument generation).

\noindent\textbf{Therapist's Strategy}   As shown in Table~\ref{tab:emolabel}, we annotated therapists' strategies based on their utterances. We have also discovered that the question is the most likely strategy employed by a therapist, especially in the early stages of counseling.

\noindent\textbf{Client's Emotion}    The Table~\ref{tab:emolabel} also shows statistics of clients' emotion labels. The distribution of user sentiment tags is relatively even. Approximately 50\% of user sentiments are various types of negative emotions (e.g., Anger, Anxiety, Fear).

\subsection{Additional Annotation Statistics}

In addition to these labels, we also annotated labels such as session-level guidance, dialog summary, turn-level step-by-step reasoning labels, and so on. Examples of these annotations can be found in Appendix \ref{sec:Example}.

\section{Evaluation}
%  ji！！！！！！！！！！！！！！！！！1
 % meteor │   self-bleu │   rouge-2 │     bleu │   rouge-1 │   data size │

\begin{table*}[ht]
  \centering
  \small
    
  \resizebox{\textwidth}{!}{
    \begin{tabular}{lcccccccccc}
    \toprule
    \multirow{2}*{ Model \& Task} & \multicolumn{7}{c}{English-Dataset} \\
    \cline{2-8} 
    &  Task Type & BertScore-P & Bleu-1 & Bleu-3 & Meteor & RougeL & Distinct-2 \\
    \hline
    Mistral-7B$_{Base}$                      & 1 & 0.824 & 0.101                  & 0.023 & 0.098 & 0.097 & 0.007\\
    Mistral-7B$_{SFT|dialog}$                       & 1 & 0.903 & \textbf{0.270} & \textbf{0.119} & 0.379         & \textbf{0.290} & \textbf{0.007}\\
    Mistral-7B$_{SFT|reaoning+dialog}$       & 2 & \textbf{0.912}& 0.266 & 0.107 & \textbf{0.397} & 0.278 & 0.001\\
    Mistral-7B$_{SFT|observation+reasoning}$ & 3 & 0.909 & 0.256                  & 0.095 & 0.383          & 0.270 & 0.001\\
    Mistral-7B$_{SFT|mix-instructions}$       & 1,2,3 & 0.877 & 0.250                  & 0.09 & 0.285           & 0.247 & 0.005\\
    \hline
    
    ChatGPT$_{Base}$                     & 1 & 0.861  & 0.163 & 0.035 & 0.193 & 0.148 & 0.001\\
    ChatGPT$_{SFT|dialog}$  & 1 & \textbf{0.888} & \textbf{0.219} & \textbf{0.071} & \textbf{0.315} & \textbf{0.216} & 0.001\\
    ChatGPT$_{SFT|reasoning+dialog}$     & 2 & 0.876 & 0.204        & 0.057         & 0.287 & 0.199 & 0.001\\
    ChatGPT$_{SFT|observation+reasoning}$ & 3 & 0.876 & 0.199        & 0.065         & 0.282 & 0.202 & \textbf{0.004}\\
    \bottomrule
    \multirow{2}*{ Model \& Task} & \multicolumn{7}{c}{Chinese-Dataset} \\
    \cline{2-8} 
    & Task Type & BertScore-P & Blue-1 & Blue-3 & Meteor & RougeL & Distinct-2 \\
    \hline
    
    GLM4-9B$_{RAG|dialog}$                     & 4  & 0.905&  0.111  & 0.025& \textbf{0.291}& \textbf{0.298} & -\\
    GLM4-9B$_{RAG|dialog+Explanation}$  & 4& \textbf{0.912}& \textbf{0.123}  & \textbf{0.029}& 0.281& 0.295 & -\\
    \hline
    \end{tabular}
    }
    \caption{Result for finetuning LLMs with different combinations of annotations. The input formats of different models can be seen in Figure 4. The mix-instruction chatbot is trained by performing instruction-tuning on multiple tasks simultaneously. }
    \label{tab:a_eval}
\end{table*}

To explore the potential of the Psy-Insight dataset, we have researched the impact of explainable annotations on the results for multiple tasks. We use the dialogue context and the annotations for finetuning (Task 3 in Table~\ref{fig: PsyInsight_example}) and RAG (Task 4 in Table~\ref{fig: PsyInsight_example}).

 % The experimental results show that the multi-task annotation and the step-by-step reasoning annotation can improve performance in mental support.
\subsection{Multi-stage Generation}

Each counseling session includes a guide $Ins$ and background $B_{i}$ and dialogue. In the $i th$ session, the conversation consists of utterances from the therapist $t_{i}$ and the client $c_{i}$. For $j th$ turns in  $i th$ session, we annotated observation $o_{j}$ for client utterance $c_{j}$, and annotated reasoning $r_{j}$ for therapist utterance $t_{j}$.

We conduct ablation experiments and treat annotation as the variable. Inspired by the chain-of-thought~\cite{wei2022COT}, we insert turn-level annotation into dialog history, to guide the model first to generate objective observations, then engage in subjective reasoning, and finally produce a response.

\begin{equation}
\label{eq2}
\begin{aligned}
Input_{i,j} &= Ins + b_{i} + \sum_{k=0}^{j-1} (o_{i,k}+r_{i,k} + t_{i,k} + c_{i,k})
\end{aligned}
\end{equation}

To assess the impact of step-by-step annotations on generation results, we employed a method similar to chain-of-thought \cite{wei2022COT}.
LLMs treat explanatory texts $o_{x}$ and $r_{x}$ as targets for the first stage of generation, and generate dialogue responses $t_{i}$ in the second stage. So, the target of the training is $ Target_{i,j}$

\begin{equation}
\label{eq3}
\begin{aligned}
 Target_{i,j} = o_{i,j}+r_{i,j}+t_{i,j}\\
\end{aligned}
\end{equation}

% $Target = t_i^j$
% $Input_i = SP+b_i+r_0+t_i^0+c_i^0+r_i^1+t_i^1+c_i^1+cdots+r_i^{j-1}+t_i^{j-1}+c_i^{j-1}$
% $Target = r_i^j+t_i^j$
% $Input_i = SP+b_i+r_0+t_0+c_0+o_0+r_i^1+t_i^1+c_i^1+o_i^1+cdots+r_i^{j-1}+t_i^{j-1}+c_i^{j-1}$
% $Target = o_i^j+r_i^j+t_i^j$

\begin{figure}[ht]
  \vspace{-0.5cm}
  \includegraphics[width=\columnwidth]{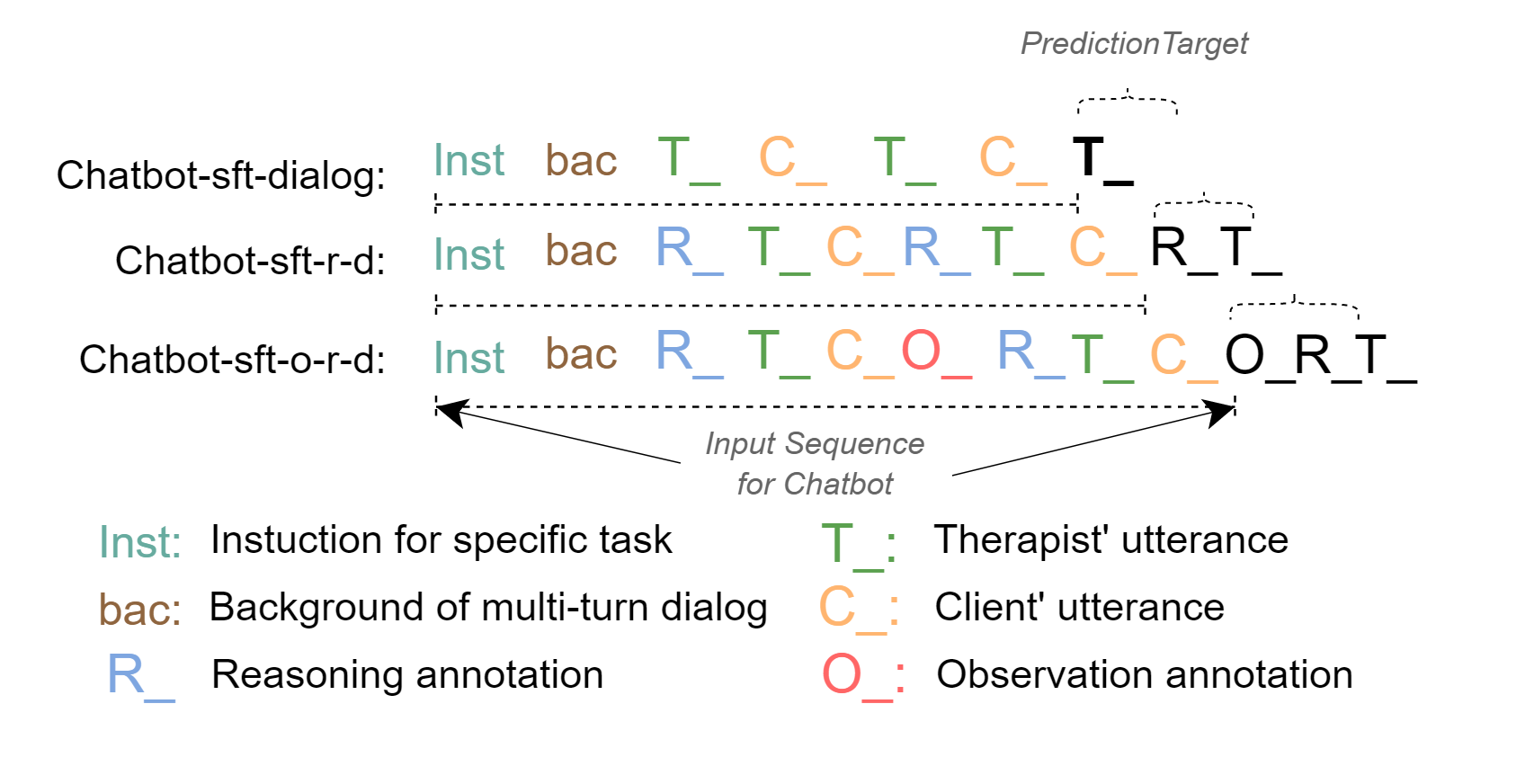}
  \vspace{-0.8cm}
  \caption{LLMs accomplish tasks by predicting the next token.For the training dataset of the Mistral7B-SFT-Reasoning model, incorporating reasoning annotations into the dialogue history and subsequently treating both reasoning and response as prediction targets allows LLMs to learn to reason first and then generate a response.}
  \label{fig:experiments}
\end{figure}
During the evaluation stage, the model first generates the annotation part and then the response. For evaluation, we only compute similarity metric between generated response $t_{i,j}$ with the ground truth response.

% \begin{figure}[h]{\linewidth}
%     \centering
%     \includegraphics[width=1\linewidth]{latex/Picture/main_picture/experiment.png}
%     \caption{Enter Caption}
%     \label{fig:experiment}
% \end{figure}

\subsection{Implementation}

We use GhatGPT-3.5-turbo and Mistral-7b-insturct-v0.2 for finetuning on the English corpus for evaluation. These models are pre-trained in the instruction dataset and suitable for learning multi-turn dialogue. And we also test ChatGLM4-9B on Chinese dataset with retrieve argument generation.

Previous studies \cite{huertaenochian2024instructionPLW} showed that Prompt Loss Weight (PLW) is suitable for learning in long multi-turn dialog. In the training stage,  we applied PLW masks technique to mask the instruction, background, and the client’s utterances. They are not involved in loss computation.

We randomly split the Psy-Insight into a training set (85\%) and a test set (15\%). For reproduction, we utilized open-source models (GLM, Mistral) for the experiment. Our code and expert evaluation results are available as open-source on our Github.

\subsection{Automatic Evaluation}

\textbf{Metrics} The automatic evaluation metrics including BertScore-P~\cite{zhang2020bertscore}, BLEU~\cite{papineni-etal-2002-bleu}, Meteor~\cite{banerjee-lavie-2005-meteor}, RougeL~\cite{lin-2004-rouge}, and Distinct-2~\cite{distinct-n}. 

% To assess the impact of step-by-step annotations on generation results, we employed a method similar to chain-of-thought \cite{wei2022COT}.
% LLMs treat explanatory texts as targets for the initial stage of generation, and generate dialogue responses in the second stage.
\begin{table}[t]
\centering
  \resizebox{0.5\textwidth}{!}{
        \begin{tabular}{lcccc}
        \hline
         \textbf{Model} & \textbf{Int.} & \textbf{Help.} & \textbf{Con.} & \textbf{Exp.} \\
        \hline
        Human & 3.44 & 3.58 & 3.78 & 3.43 \\
        GLM4-9B & 2.90 & 2.76 & 2.79 & 3.21 \\
        GLM4-9B$_{reasoning}$ & 3.12 & 2.88 & 2.85 & 3.25 \\
        GLM4-9$_{obs+reasoning}$ & 3.00 & 2.62 & 2.56 & 2.78 \\
        \hline
        \end{tabular}
    }
\caption{Human interactive evaluation focuses on interactivity, helpfulness, comfort, and explainability.}
\label{tab:human_eval}
\end{table}

\noindent\textbf{Result} Table ~\ref{tab:a_eval} shows the automated metrics results of the instruction-tuning. Compared with the baseline, all finetuned models show improvement in automatic metrics. This suggests that Psy-insight is helpful for the model to better generate responses for mental support. Compared to pure dialog fine-tuning, incorporating step-by-step annotations results in increased metrics such as BertScore-P, Meteor, and Distinct-2 scores.

\subsection{Human Evaluation}

We designed two types of human evaluation experiments. 

\noindent\textbf{Human Interactive Evaluation} As shown in Table~\ref{tab:human_eval}, we conducted an ablation experiment with the Psy-Insight dataset to investigate the generation quality of ChatGLM4 with dialog and annotation. We used different labels from Psy-Insight to finetune ChatGLM4-9B models and compared their generated results with the original human counseling dialogue.

\noindent\textbf{Similar Cases Evaluation} As shown in Table \ref{tab:dataset-compare}, to compare the counseling quality across datasets, we randomly selected English cases from the Psy-Insight and compared them with similar cases from the Esconv \cite{liu2021Esconv} English dataset. In the same way, we compared Psy-Insight's Chinese cases with similar cases from the synthetic Smile \cite{qiu2023smile} dataset. 

\noindent\textbf{Metrics} We recruited 10 student volunteers and 5 experts for human evaluation on 60 random selected responses. They were asked to compare the responses based on the following metrics: \noindent\textbf{Interactivity}: Does the therapist have the intention to continue the conversation? \textbf{Helpfulness}: Whether the suggestion suitable for client? \textbf{Comforting}: Whether the response is useful for comforting client? \textbf{Explainability}: Is the self-explanation by the model reasonable in counseling?  We standardized the evaluation results to ensure consistency among raters.
\begin{table}[t]
  \centering
  \small
    
  \resizebox{0.5\textwidth}{!}{
    \begin{tabular}{ccccc}
    \toprule
    \multirow{2}*{ Psy-Insight vs.} & \multicolumn{2}{c}{Esconv\citeyearpar{liu2021Esconv}} & \multicolumn{2}{c}{Smile\citeyearpar{qiu2023smile}}\\
    
    & Win & Lose & Win & Lose \\
    \hline
    Interactivity & 55 & 21& 57 & 17 \\
    Helpfulness & 44& 28 & 52 & 24 \\
    Comforting & 56 & 27& 46 & 34 \\
    Explainability & 45 & 29 & 42 & 30 \\
    \hline
    Overall & 50 & 32 & 51 & 25 \\
    \hline
    \end{tabular}
    }
    \caption{Result of the expert A/B test evaluation among Psy-Insignt and other baseline dataset. Experts not only provided scores but also offered comparative assessments for similar cases, as detailed in Appendix \ref{sec:comparison}. }
    \label{tab:dataset-compare}
\end{table}
In case comparison, we ensured randomness and similarity by clustering cases's topics with sentence-Bert~\cite{2019-sentence-bert}, selecting the top 5 detailed cases per major category, and assessing content/theme similarity with ChatGLM4-9b. We then chose 20 pairs for expert scoring (Table ~\ref{tab:dataset-compare}) and 10 for comparative assessments (Appendix  ~\ref{sec:comparison}).
% Examples of these responses and the explanations given to the assessors can be found in Appendix \ref{sec:Guideline}.

\noindent\textbf{Result} As shown in Table~\ref{tab:human_eval}, LLMs trained with reasoning annotation generates high-quality counseling conversations. However, there is still a remarkable gap between the mental support LLMs and professional therapists. Table~\ref{tab:dataset-compare} shows that Psy-Insight achieves better quality than Esconv in English dialogue data, and its Chinese dialogue quality surpasses that of the synthetic dataset Smile. 

% For the method of similarity judgment and more comparison samples, please refer to Appendix \ref{sec:comparison}.

% & 3.44 & 3.58 & 3.78 & 3.53 \\
% & 2.90 & 2.76 & 2.79 & 3.25 \\
% & 3.51 & 3.38 & 3.40 & 3.42 \\
% & 3.00 & 2.62 & 2.56 & 2.78 \\

\section{Conclusion and Future Work}

We provide Psy-Insight, a high-quality bilingual psychological counseling corpus, and annotated it with step-by-step reasoning and multi-task labels. Rich annotations make the Psy-Insight dataset suitable for multi-task learning in mental support. Additionally, with step-by-step explanations, LLMs can understand the reasoning behind counseling. 

In future work, we plan to utilize expert assessments in Appendix~\ref{sec:comparison}. These comments from professional counselors are high-quality human feedback. We hope to utilize them with techniques such as reinforcement learning or DPO~\cite{2024DPO}, enabling LLMs to counsel based on human preferences.
% Furthermore, with specific instructions for multiple tasks, Psy-Insight is a perfect dataset for the LLM’s instruction-tuning.
% In future work, we plan to evaluate the performance of  mental support LLMs on DPO~\cite{2024DPO}, utilizing expert assessments in appendix F.

% \subsection{Appendices}

\clearpage
\section*{Limitations}
Although the Psy-Insight dataset offers rich psychological annotation for at least 5 NLP tasks, we only evaluate the instruction tuning in English dialogue due to the limited resources. We hope the statistical data we provide can help future research.

\subsection*{Copyright}
\label{sec:copyright}
We collect datasets from crawled blogs and books and also extract conversation from raw common crawl datasets including the book3 and Massive Never-ending BT Vast Chinese corpus project. All the copyright information and data sources are recorded in our GitHub repository.

\bibliography{custom}
\clearpage
\appendix

% Acknowledge efforts, ask open-ended, reflect feelings, respect boundaries, encourage sharing.

\section{Ethical Evaluation}
\label{sec:ethical}
\subsection*{Privacy}

Considering the potential ethical risks. We filter the privacy information in Psy-Insight to ensure that our data is only related to the cases and not to personal identities. The raw crawled data contains sensitive data like names, URL links, contact information phone numbers so on. We put a lot of effort in the anonymization process.
 
For example, as the title of both sides in the consultation, real names are frequently collected in counseling data. In our experiment, we observe that masking the real name directly will cause a performance decrease in model finetuning. As a result, We take the rule-based method to replace the title in the dialogue content with pronouns and annotate common names based on gender and family role dialogues, such as "therapist", "client", and "Smith" for adult men, and "Mary" for adult women, and so on. Similarly, we use roles-based tools to filter sensitive personal information in raw data.  We guarantee that our Psy-Insight does not include any personal privacy information. All names in Psy-Insight are aliases.

\subsection*{Ethical Risk}

We safeguard the anonymity of the data. In our repository, users must agree to a statement that they cannot trace or de-anonymize the content of the consultation data. Our dataset is intended solely for academic research.

Beyond privacy concerns, we are also aware of the high risks associated with training LLMs to provide psychological counseling services. To prevent models trained on Psy-Insight from generating misleading responses, we have cleaned and risk-assessed the counseling dialogues in the Psy-Insight dataset. First, we filtered out some counselings lacking psychotherapy labels from the crawled dataset. We cannot ensure the controllability and scientific nature of dialogues with unclear data sources. Next, we separated counseling dialogues with negative emotional (eg,. anxiety) and behavioral keywords based on emotional tags, which may include actions such as suicide and depression. Finally, we had three professional psychological therapists assess the dialogue content and filter the useless responses. All of the psycho-experts in our work have at least national second-level psychological certificates. Our effort aims to ensure that the counseling dialogues in the Psy-Insight dataset align with psychological standards and ethics.

% Our data annotation work has been approved by the University's Academic Ethics Committee.
% 

\section{Reproduction}
\label{sec:reproduction}
In this section, we provide an overview of our experimental setting in Evaluation. All experiments were conducted on 2 NVIDIA RTX 3090 GPUs. For finetuning, we apply \cite{hu2021lora} Lora for instruction-tuning LLM. We set alpha at 128 and rank at 256.

We finetune the English part of Psy-Insight with a batch size of 8 and a learning rate of 0.00002. For equality in evaluation, all models are trained for 220 epochs. We take the transformers-4.36.2 \cite{wolf-etal-2020-transformers} and peft-0.9.1 frameworks for training.  

% *
% \onecolumn
% *
\onecolumn
\section{Construction Workflow Example}
\label{sec:DataConstruction}
% \begin{figure}[H]
%     \centering
%     \includegraphics[width=0.6\textwidth]{latex/Picture/main_picture/Workflow.png}
%     \centering
%     \caption{The construction workflow of Psy-Insight Dataset.}
%     \label{fig:data-workflow}
% \end{figure}
\begin{figure}[H]
    \centering
    \includegraphics[width=0.95\textwidth]{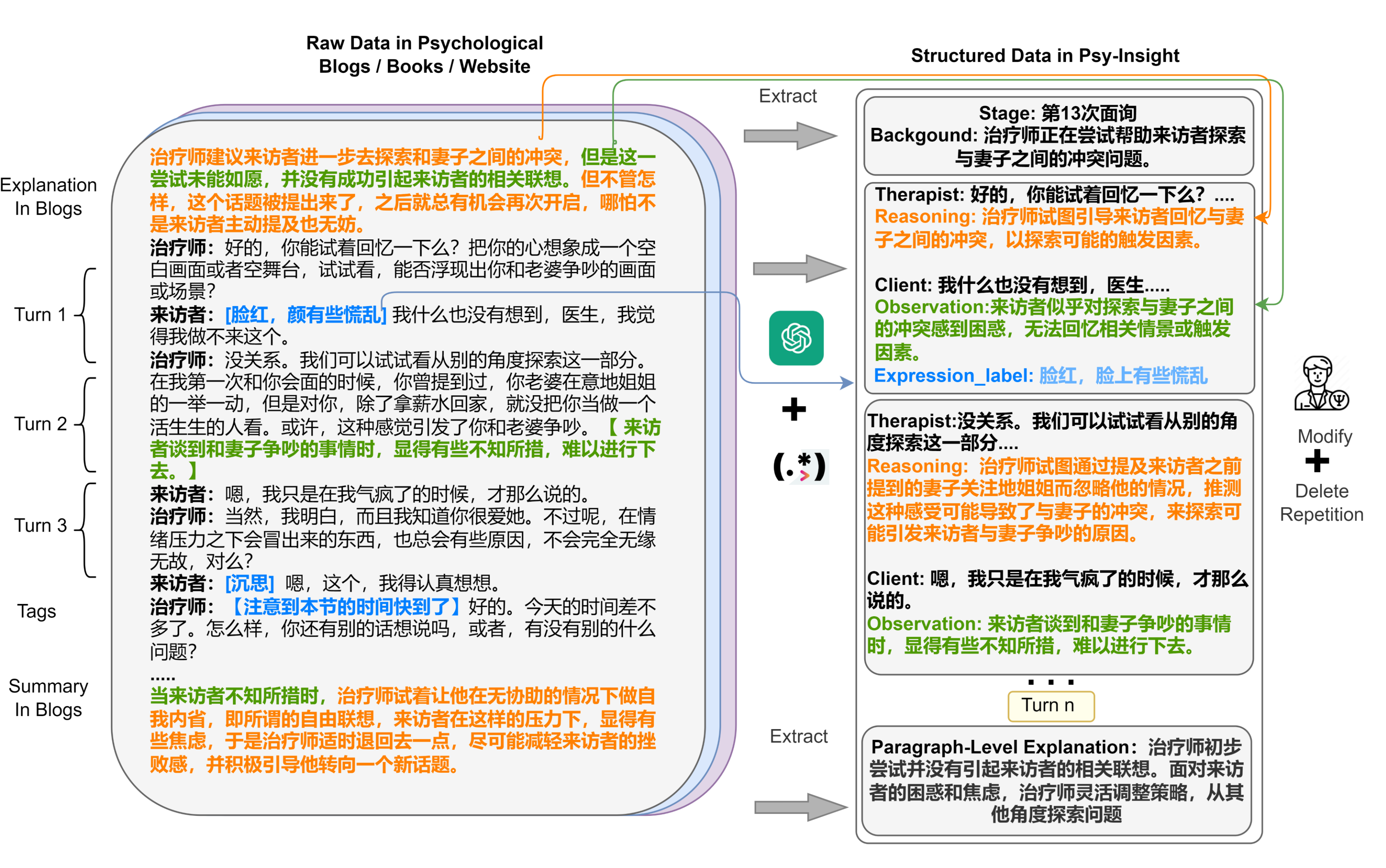}
    \centering
    \caption{The data collection process in Psy-Insight Dataset. We use psychological labels as anchors to locate dialogue-like structures from raw web data. After positioning, we structure the dialogue data and descriptive text using regular expressions and LLMs. In the workflow of dialog collection, GLM4-9B and Mistral7B are used for formatting rather than synthesizing them from scratch. Non-dialog labels are extracted from explanatory paragraph in raw textbooks. Furthermore, a long-context LLM agent is used to map these structured labels into related dialog turn.}
    \label{fig:data-construction}
\end{figure}
% \twocolumn
\begin{multicols}{2}
% \lipsum[2-3]

% The data collection process in Psy-Insight Dataset. We vaguely locate the psychotherapy in raw data, an extract structured annotations for Psy-Insight. In workflow of dialog collection, GLM4-9B and Mistral7B are used for mapping  rather than synthesizing them from scratch. 

% The data collection process in Psy-Insight Dataset. We vaguely locate the psychotherapy in raw data, an extract structured annotations for Psy-Insight. ChatGPT is only used for mapping and extraction. Human annotators will evaluation the extraction result from GPT and modify the annotation.
\end{multicols}

% ... 
% \begin{figure}[H] 
% foo 
% \end{figure}

% \begin{minipage}{\textwidth}
%     \centering
%    \includegraphics[width=1\linewidth]{latex/Picture/main_picture/Data_Colleciton.png}
%    Figure 6: The overall data collection process in Psy-Insight Dataset. We vaguely locate the psychotherapy in raw data, an extract structured annotations for Psy-Insight. ChatGPT is only used for mapping and extraction. Human annotators will evaluation the extraction result from GPT and modify the annotation.
%     \caption{The overall data collection process in Psy-Insight Dataset. We extract in-context. }
%     \label{fig:enter-label}
% \end{minipage}

% \begin{figure*}[H]
%     \centering
%    \includegraphics[width=0.8\linewidth]{latex/Picture/main_picture/Data_Colleciton.png}
%     \caption{The overall data collection process in Psy-Insight Dataset. We extract in-context. The }
%     \label{fig:enter-label}
% \end{figure*}
% d

% \twocolumn
% \onecolumn
% \clearpage

\section{Detailed Statistics}
\begin{table}[ht]
\small
\centering
\resizebox{\textwidth}{!}{
            \begin{tabular}{
            >{\centering\arraybackslash}p{3cm}>
            {\centering\arraybackslash}c>
            {\centering\arraybackslash}cc>
            {\centering\arraybackslash}r>
            {\centering\arraybackslash}c>
            {\centering\arraybackslash}cp{4cm}} % 将 p 列类型全部改为 m，并在前面添加 \centering\arraybackslash
            \toprule
        \textbf{Dataset} & \multicolumn{1}{c}{\textbf{Domain}} & \textbf{Dialog Source} &\textbf{Avg. Turn} & \textbf{Total Turn} & \textbf{Language} & \textbf{Annotations}\\
        \midrule
        Empathetic Dialogue  & Emotional & forum & 4.31 & 24,850 & en & Emotion \\
        \citeyearpar{rashkin2018towards} & Support &  &  &  &  &  \\
        ESConv                      & Emotional & crowd & 29.8 & 1,053 & en & Background,Strategy \\
        \citeyearpar{liu2021towards} & Support & workers &  &  &  & Emotion,Instensity \\
        MedDialog  & Medical & hospital & 3.19 & 1,145,231 & en
        & Depression \\
        \citeyearpar{zeng2020meddialog} &  Dialogue & & & 257,454 & zh \\
        D4  & Depression & hospital & 21.6 & 1,339 & zh & Depression \\
        \citeyearpar{yao2022d4} & Diagnosis & & &  &  &  \\
        PsyQA & Mental & forum & 2.51 & 22,346 & zh & Background,Strategy \\
         \citeyearpar{sun2021psyqa} & Health &  &  &  &  & Emotion,Instensity \\
        Smile & Mental & ChatGPT & 10.4 & 55,165 & zh & - \\
        \citeyearpar{qiu2023smile} & Health &  &  &  &  &  \\
        SoulChat                       & Mental & ChatGPT & 20 & 2,300,248 & zh & - \\
        \citeyearpar{chen2023soulchat} & Health &   &  &  &  &  \\
        \midrule
        \multirow{2}*{\textbf{Psy-Insight}} & Mental & book & 46
        & 6,208 & en & 
        Explainable Annotations
        
         \\
         & Health &  blog & 77 & 
        5,776 & zh & 
        Multi-task Labels
         \\
        \bottomrule
    \end{tabular}

    }
    % \vspace{-0.7cm}
    \vspace{-0.3cm}
\caption{Comparison of the Psy-Insight dataset with baseline datasets. Compared to previous datasets, Psy-Insight includes face-to-face counseling from real life, featuring longer average dialogue turns (46 in English, 77 in Chinese). Unlike previous datasets that primarily focus on short labels and single subtasks, Psy-Insight annotates dialogues at various granularities with step-by-step explanatory and rich multi-task labels.}\label{mosi}
% As shown in Table~\ref{tab}, we compared the key information in the Psy-Insight dataset with that of previous datasets.
\label{tab:compare_with_datasets} % 添加标签

\end{table}

\label{sec:Statistics}
\begin{figure}[ht]
    \centering
    \begin{minipage}{\textwidth}
        \centering
        \begin{subfigure}[b]{0.45\textwidth}
            \includegraphics[width=\linewidth]{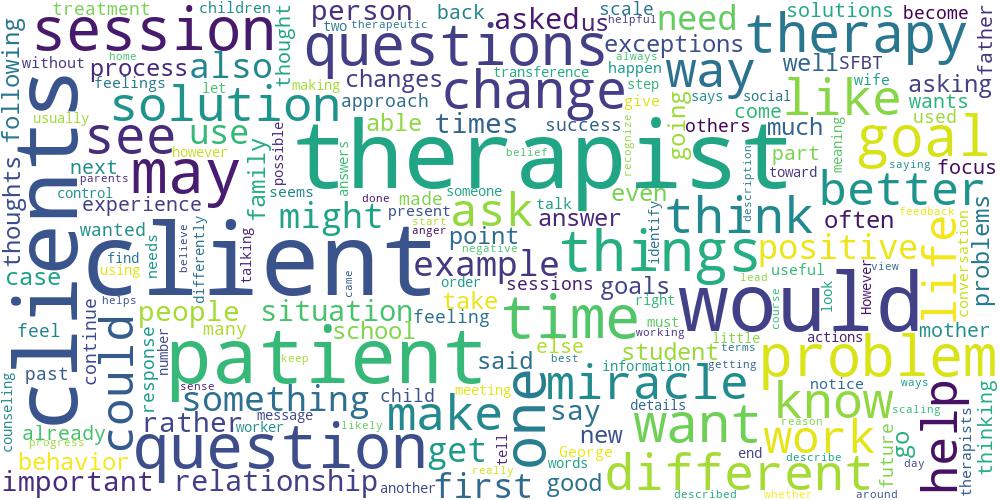}
            \caption{Word clouds of English explainable annotations in Psy-Insight.}
            \label{fig:sub3}
        \end{subfigure}
        \hfill
        \begin{subfigure}[b]{0.45\textwidth}
            \includegraphics[width=\linewidth]{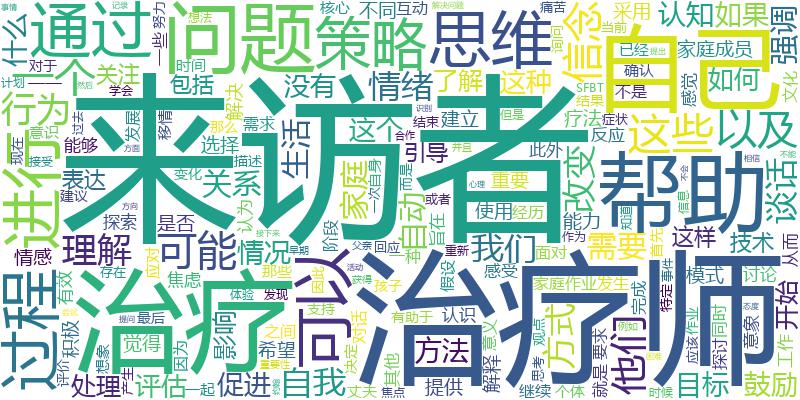}
            \caption{Word clouds of Chinese explainable annotations in Psy-Insight.}
            \label{fig:sub4}
        \end{subfigure}
        \caption{Common words in annotations.}
       
    \end{minipage}
    \par\bigskip % add some vertical space
    \begin{minipage}{\textwidth}
        \centering
        \begin{subfigure}[b]{0.45\textwidth}
            \includegraphics[width=\linewidth]{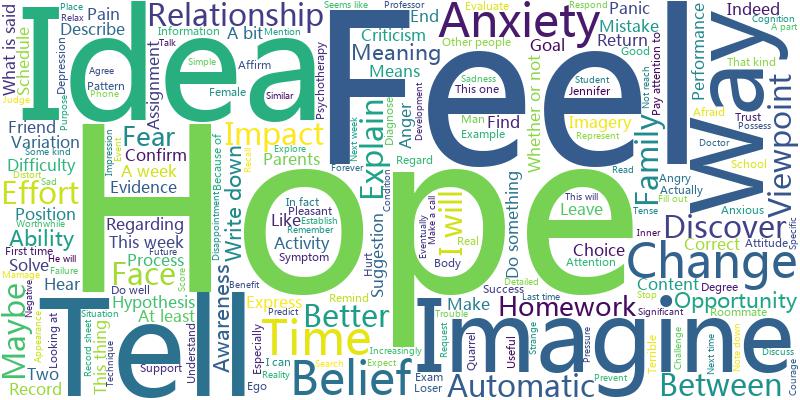}
            \caption{The translation of previous Chinese explainable annotation dialog.}
            \label{fig:sub1}
        \end{subfigure}
        \hfill
        \begin{subfigure}[b]{0.45\textwidth}
            \includegraphics[width=\linewidth]{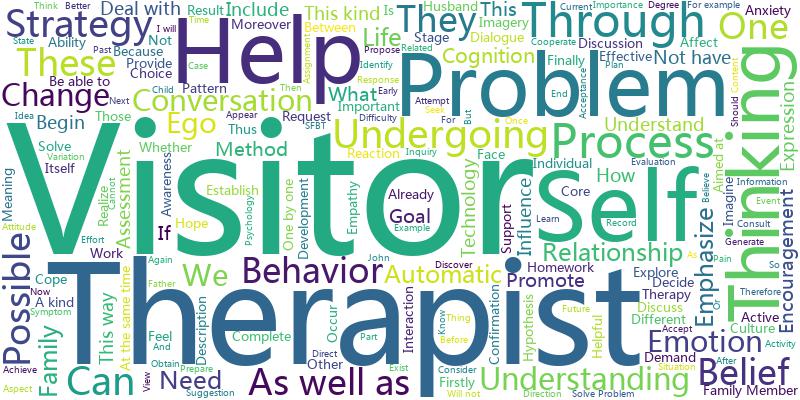}
            \caption{The translation of Chinese reasoning picture.}
            \label{fig:sub2}
        \end{subfigure}
        \caption{Translation for previous Chinese word picture.}
        % \label{fig:sub2}
    \end{minipage}
    % \caption{Word cloud figures of English annotation and Chinese annotation. The following two pictures show the corresponding translation information in the previous Chinese data.}
    \label{fig:word_cloud_figures}
\end{figure}

\begin{figure}[H]
    \centering
    \includegraphics[width=0.775\textwidth]{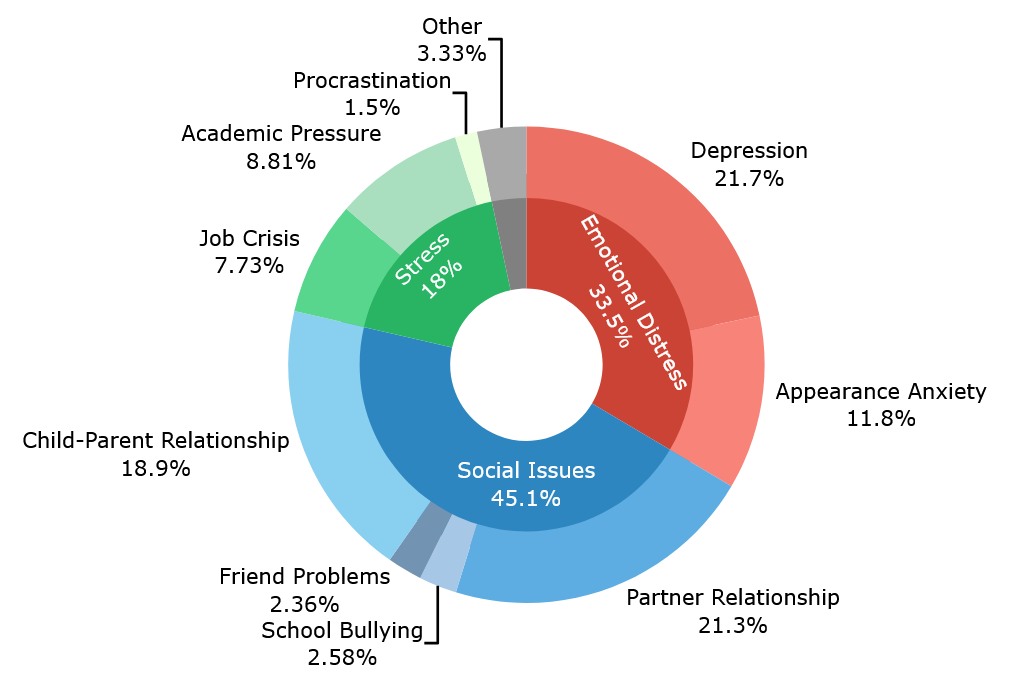}
    \centering
    \caption{Statistics of topics in counseling of Psy-Insight.}
    \label{fig:enter-label}
\end{figure}

\clearpage

\section{Data Examples of Psy-Insight}
\label{sec:Example}
% \vspace{-14.0em}
% \vspace{-5.0em}
% \begin{minipage}[H]{.5\linewidth}
%     \centering
%     dlfsakjdlkaf
% \end{minipage}
\begin{table}[H]

\centering
\small
\resizebox{\textwidth}{!}{
    \begin{tabular}{p{6cm}p{1.7cm}p{3.5cm}}
     \hline
    \textbf{Topic} & \textbf{Session ID} & \textbf{Psychotherapy} \\
    \hline
    Addressing academic challenges in students. & 32  &  Solution Focused  Brief Therapy (SFBT)\\
    \hline
    \textbf{Background} &  \multicolumn{2}{l}{\textbf{Guide}} \\
    \hline
    A student struggling with attendance and academic performance, as discussed in the previous session. & \multicolumn{2}{p{5.2cm}}{Acknowledge efforts, ask open-ended, reflect feelings, respect boundaries, and encourage sharing.} \\
    \hline
    \multirow{2}*{\textbf{Dialog}} & \textbf{Strategy 
     /} & \textbf{Reasoning /} \\
     & \textbf{Emo-label} & \textbf{Observation} \\
    \hline
        \textbf{Therapist}: Yes, you’ve been able to get to school sometimes but not as much as you may want. I’m impressed, though, with how you are at least trying to make it to school these last few weeks since we talked. There must be a lot going on that makes it harder for you to come to school as much as you would like. I’m curious; is there anything that would be helpful to talk about that might help even a little bit? & Question & Therapist acknowledges the student's efforts and explores potential topics for discussion to offer support.\\
        \textbf{Beth}: Maybe … I don’t know if I can … not sure …it is very hard and I don’t know what will happen if I talk about it. & Anxiety, Fear & Beth expresses uncertainty and difficulty in opening up.\\
        \textbf{Therapist}: Well, in what way do you think it might help if you did share it with me or someone else? Do you think it would make things better for you … like feeling like coming to school? & Question & Therapist encourages Beth to consider the potential benefits of sharing her concerns.\\
        \textbf{Beth}: I don’t know. It might make things even worse. & Fear & Beth fears that sharing could worsen things\\
        \textbf{Therapist}: It must be very important to you if it might make things worse. Even though it sounds like if you were to get some help with this issue, it might make your life easier in doing the things you seem to want to do, like school. That is a tough place to be. What would be the most helpful for us to do that might help move you to a better place and make you feel better? & Question, Reflection & Therapist acknowledges the significance of Beth's concerns and explores ways to support her in improving her situation.\\
    \hline
     \multicolumn{3}{l}{\textbf{Summary}} \\
    \hline
      \multicolumn{3}{p{11.2cm}}{The student begins to share her experience, possibly due to feeling accepted and supported by the therapist, indicating a growing trust in the therapeutic relationship.} \\
    \hline
    \end{tabular}
}
\caption{An example session in the English Dataset of Psy-Insight.}
\label{tab:english_example}
\end{table}

\begin{multicols}{2}
% \lipsum[2-3]

The data in the Psy-Insight dataset is collected and annotated in session units. Table~\ref{tab:english_example} and Table~\ref{tab:chinese_example} show the data units for English and Chinese data, respectively. Each session contains 5 to 50 rounds of dialogue between a counselor and a client on a specific topic. Our annotation labels include Session-Level and Turn-Level labels. Session-Level labels include background, guidance, session topic, and counseling summary. Turn-Level labels are annotations for each turn of dialogue in multi-turn conversations. We labeled different aspects of the counselor's and client's dialogue content. For the counselor's dialogue, we annotated counseling strategies and subjective reasoning. For the client's dialogue, we annotated emotional classification results and observations of dialogue facts from the counselor's perspective. Due to space limitations, more Chinese dialogue examples and English dialogue cases can be found on our github website.

\end{multicols}
\clearpage
% 注 resize之后 的表格很奇怪，可以把表格向上挤压
\begin{CJK*}{UTF8}{gbsn}
\begin{table}[H]

\centering
\small
\resizebox{\textwidth}{!}{
    \begin{tabular}{p{8cm}p{1.7cm}p{3cm}}
     \hline
    \textbf{Stage} & \textbf{Session ID} & \textbf{Psychotherapy} \\
    \hline
    The 4th Session  & 23  &  Postmodern Therapy\\
    \hline
    \textbf{Background} &  \multicolumn{2}{l}{\textbf{Guide}} \\
    \hline
    来访者是一个年轻女性，面临与家庭和亲密关系相关的挑战，希望在情感认知和复杂性方面得到支持和理解。可能正在寻求解决家庭和婚姻中的困境，希望更清晰地了解自己的情感和选择。 & \multicolumn{2}{p{5.2cm}}{治疗师需要通过逐步询问，帮助她分离问题，减轻负面情绪并澄清问题影响，以提升她的投入感和生活应对能力。} \\
    \hline
    \multirow{2}*{\textbf{Dialog}} & \textbf{Strategy 
     /} & \textbf{Reasoning /} \\
     & \textbf{Emo-label} & \textbf{Observation} \\
    \hline
        \underline{治疗师}: 你愿意挑战自己的恐惧并更加开放地表达自己，但是你还是希望一点一点慢慢来，是吗？ & Question & 治疗师通过逐步询问引导来访者澄清问题。 \\
        \underline{来访者}: 当然。 & Neutral & None\\
        \underline{治疗师}: 来访者，你从哪里学到如何做一名女性、妻子、母亲的？ & Question &  通过提问探索来访者的性别角色认知。\\
        \underline{来访者}: 我不知道。我从来没有想过这个问题。 & Neutral & 来访者对这个主题缺乏反思。\\
        \underline{治疗师}: 嗯，我想知道你从哪里学到要照顾他人，将自己的需要摆在他人需要的后面，有时甚至应该牺牲小我成全他人的？ & Question & 进一步询问来访者自我牺牲和他人优先的观念来源。\\
        \underline{来访者}: （现在依然如此）我想我是从父母那里学来的，还有我父亲对待我母亲的方式。 & Others & 来访者认识到她的行为模式和观念可能来自父母的影响。\\
        \underline{治疗师}: 你不认为你现在的生活方式就是在学习你母亲的做法吗——关于如何做一名女性、妻子和母亲？ & Question & 让来访者反思她是否在重复母亲的生活模式。\\
        \underline{来访者}: 我不确定，我从没想过我是否在重复我母亲的生活方式。我知道她从未想过要出去工作或去读书。从这个角度来看，我和她有很大的不同。 & Neutral & 来访者表示不确定，也指出她们之间的显著差异。\\
        \underline{治疗师}: 是的，这看起来的确是个真正的差异。我想知道你能否找到你和她一致的地方？ & Question & 治疗师承认差异，并鼓励来访者寻找和母亲生活方式相似的部分。\\
        \underline{来访者}: 嗯，我认为我母亲十分传统。在她看来，男人是整个家庭的领导，是家庭收入的主要来源，如果你希望的话，他也可以成为家庭的保护者。而女性的工作就是养育孩子、照料整个家庭，我猜还包括照顾丈夫。 & Neutral & 来访者描述她母亲的传统观念和性别角色分工。\\
        \underline{治疗师}: 在你母亲教授给你的诸多观念中，你接纳了其中的多少？ & Question & 治疗师询问来访者对母亲观念的内化程度。\\
       \underline{来访者}: 嗯，我猜在我结婚的前15年中，我一直在步母亲的后尘。实质上我在我的婚姻生活中一直在做她那样的女性。现在，这成为了我的问题。我觉得我不希望再这样下去了。我不希望以牺牲我的利益为代价去担负起让来访者幸福的责任。我不希望让我的生活继续围着来访者转。但是我又痛恨自己的这些想法。我为自己想要追求自己的职业而感到内疚。 & Guilty & 来访者反思自己在婚姻中的行为模式，并表达出对改变的渴望和内心的矛盾与内疚感。\\

    \hline
     \multicolumn{3}{l}{\textbf{Summary}} \\
    \hline
      \multicolumn{3}{p{11.2cm}}{作为咨询师，我应当引导来访者质疑、反思并逐步替换这些不健康的文化信念，建立更积极的自我认知.} \\
    \hline
    \end{tabular}
}
\caption{A data unit in the Chinese part of Psy-Insight. Due to the page limitation, we only show half part of dialog in this 19-turn dialog. Table~\ref{tab:translation} shows its English translation.}
\label{tab:chinese_example}

\end{table}
\end{CJK*}

    \vspace{-0.2cm}
\section{Expert Assessments of Psy-Insight and Baseline Datasets}
\label{sec:comparison}

\begin{table}[H]
    \centering
    \small
    \resizebox{\textwidth}{!}{
        \begin{tabular}{p{6.5cm}p{6.5cm}}
        \hline
        \multicolumn{2}{l}{\textbf{English Counseling Case}}\\
         \hline
        \textbf{Psy-Insight} & \textbf{Esconv \cite{liu2021Esconv}} \\
        \hline
        \multicolumn{2}{l}{\textbf{Source}}\\
        
        \hline
         Textbook\& Psychological blogs & Crowdworkers \& Volunteers  \\
        %  \hline
        %  \multicolumn{2}{l}{\textbf{Topic}}\\
        % \hline
        % \multicolumn{2}{l}{Depression}\\
        \hline
        % \begin{tabular}{p{6.5cm} p{6.5cm}}
\begin{minipage}[t]{6.5cm}
\underline{seeker}: Well, I certainly would be a lot more cheerful. \\
\underline{supporter}: Mmm-hmm. Would that … do you think that would have some effect on this getting better business? \\
\underline{seeker}: Probably. \\
\underline{supporter}: Yeah. Oh. Okay. That was the only question to pop into my head while I was …  Do you have any? All right. I think we should take a break and we’ll be back in 10 minutes. Okay. Just relax. \\
\underline{seeker}: Yes, we do. \\
\underline{supporter}: Okay. You need more practice in just relaxing. Just … \\
\underline{supporter}: Well, I’m glad you both came today and I thought I saw some new talent that I hadn’t heard about before today. \\
\underline{supporter}: I wrote down some of the stuff the team said about you. \\
\underline{seeker}: Yes. \\
\underline{supporter}: And they are very impressed by how concerned you are and how caring you are and your perseverance. And you—at the same time being smart, taking one day at a time. But also, I mean you have been seeking help for four years. \\
\underline{seeker}: Yes. \\
\underline{supporter}: And you never gave up. And it must— \\
\underline{seeker}: I couldn’t. \\
\underline{supporter}: Oh, sure, but it also must take some knowing about what is possible to sort of give you the power to continue to search knowing that you could. \\
\underline{seeker}: Yes. \\
\underline{supporter}: Not believing that it’s impossible. There’s force in that. And at the same time being cautious at this point is a sign of wisdom. Okay. And someone pointed to the fact that you’re very, very good at sharing your happiness about him being better and showing him that. \\
\underline{seeker}: Yes, but I think that’s … that’s one of …  that you have pointed out and you have tried … to tell me to do that sometimes when I just feel: “What shall I do, what can I do, how can I do?” And I think of what you are simply saying during these meetings we have had in 1½ years. And you have said, “Try to see the good things that were happening just now. And, my older sons, I think we are very close, and we are talking, and they are seeing when I am not feeling so well about everything and they try to support me there. \\
\textbf{... turn X 10} \\
% \underline{seeker}: Support you there. \\
% \underline{seeker}: [Laughs.] \\
% \underline{supporter}: That’s good. That’s very good. That’s very good. \\
% \underline{supporter}: Yeah. Yeah. \\
% \underline{seeker}: [Laughs.] \\
% \underline{supporter}: That’s good. I’m glad to hear that. And, they wanted to say … wanted me to say about how impressed they have been also with your outstanding English. \\
% \underline{seeker}: Thank you. \\
% \underline{supporter}: So … I think it’s, you know … outstanding is a good word for it. Um. I was also impressed with your ability to observe things and your sense of humor. \\
% \underline{seeker}: It’s pretty cheap, but it works. \\
% \underline{supporter}: Yeah. Sure. It, you know … I think it would be a good idea to make sure that good things happen to you so that you can make … continue to make things better. I don’t know quite how you do that, you know, but, because, you know … obviously from what you were saying, I agree with that. That, you know, when good things happen you feel better. Um. I was thinking that, you know, that when you wake up in the morning and you have good feelings in your stomach, you should try to make friends with them. \\
\end{minipage} &
\begin{minipage}[t]{6.5cm}
\underline{seeker}: Hello there, would you be able to give me some advice? \\
\underline{supporter}: I would be glad to. What is troubling you? \\
\underline{seeker}: The first lockdown was bad enough but the second has really thrown me into despair. I simply cannot cope without being able to see friends and family for much longer. \\
\underline{supporter}: I completely understand and had similar experiences. What were some ways you tried to interact with people last shut down? \\
\underline{seeker}: I use facebook quite a lot and it helps but it cannot compensate for face to face contact. I really need to be able to see people. \\
\underline{supporter}: Digital contact does have its limitations. Could you meet with people outside at a distance? \\
\underline{seeker}: Well, part of the problem with that is that I am not sure what the rules are any more! We have been told so many different things that I am out of the loop. \\
\underline{supporter}: I understand. Try sticking to the basics to make it less stressful. Wear a mask outside at a distance. \\
\underline{seeker}: I always do wear a mask.. I think we legally have to :). I really miss little things like going out for coffee or to the gym though. It makes me feel like will never be the same again. \\
\underline{supporter}: Have to talked about this with friends? They may be feeling the same way. Just talking this out with the may be reassuring. \\
\underline{seeker}: I have a little bit but I don't want to burden them. The other aspect is that it is making me anxious about many others things too. I keep having morbid thoughts about what will happen if the pandemic never ends. \\
\underline{supporter}: That’s understandable. Maybe talking to a professional who you don’t have to worry about burdening with your concerns will  allow you to talk about your concerns. \\
\underline{seeker}: That's why I'm here ;). You are being very helpful, thank you! It's just nice to be able to tell people what I am worried about and not have them think that I am just being silly. \\
\underline{supporter}: Many people are struggling with the current state of affairs. Don’t be afraid to talk to people they may also be afraid to talk to people and you could miss out on supporting each other. \\
\underline{seeker}: That does sound like really good advice. The other thing I haven't mentioned yet is that I have been having financial pressures because I am on furlough. I am worried that if this lasts much longer I will end up deep in debt. \\
% \textbf{... turn X 6} \\
% \underline{supporter}: I am sorry that wasn’t helpful. \\
% \underline{supporter}: Have you tried looking into local recourses in your city? \\
% \underline{seeker}: I felt that you were very helpful. It is nice to just know that someone is listening. I do not think that there is much in the way of local resources at the moment - everyone is very stretched. \\
% \underline{supporter}: That is true it has been hard for many people. \\
% \underline{seeker}: I do feel better for having been able to talk to you about this - thank you very much! \\
% \underline{seeker}: I think I will be able to carry on for a bit now. Thanks again and best wishes! \\
% \underline{supporter}: I’m glad I could provide you some support. \\
\end{minipage} \\
% \end{tabular}
 \hline
        
        \end{tabular}
    }
    \vspace{-0.2cm}
    
    \caption{The case comparison of Psy-insight and Esconv on the similar topic: Distress.}
    \label{tab:Psy-insight-vs-smile}

    \end{table}

\begin{table}[H]
    \begin{CJK*}{UTF8}{gbsn}
    \centering
    \small
    \resizebox{\textwidth}{!}{
        \begin{tabular}{p{6.5cm}p{6.5cm}}
         \hline
        \textbf{Psy-Insight} & \textbf{Esconv\cite{liu2021Esconv}} \\
                 \hline
        \multicolumn{2}{l}{\textbf{Dialog ID}} \\
        \hline
        \textbf{1} & \textbf{2} \\
        \hline
         \multicolumn{2}{l}{\textbf{Experts’ Assessments}} \\
        \hline
\multicolumn{2}{p{13cm}}{\underline{专业咨询师A}：1更像真实的咨询，双方有情感流动，有非言语信息。2比较像机器人，一直在回答来访者的问题，情感互动较弱，对求助者的情绪关注不够。}\\ %第一行
\multicolumn{2}{p{13cm}}{\underline{专业咨询师B}：1非常尊重求助者的决定，给出建议让求助者自己做决定；2依然更关注问题解决。咨询师不应替求助者做决定。}\\ %第二行
\multicolumn{2}{p{13cm}}{\underline{心理系学生C}：对话1更好，有共情和尝试让来访者提出自己曾经尝试过的有效方法。}\\ %第三行
\multicolumn{2}{p{13cm}}{\underline{心理系学生D}：对话2更好，提出了具体的解决办法；对话1的处理方式可能并不太合适。}\\ %第四行
\multicolumn{2}{p{13cm}}{\underline{心理系学生E}：1有一种逼着seeker做决定的感觉，这让我觉得不太好。因此尽管1提出的解决方案可能有效，也未必真的就起到作用，还可能产生二次伤害。和其他的2相比，感觉这个2在安慰有效性上比较欠缺，也是说了一些“正确的废话”。}\\ %第五行
\multicolumn{2}{p{13cm}}{\underline{心理系学生F}：对话1更好，在敏感话题给予了隐私维护。}\\ %第六行
\multicolumn{2}{p{13cm}}{\underline{心理系学生G}：对话1更好，对话1中清楚询问了情况并给出了具体的建议，并安抚了求助者的情绪。对话2则更像机器人的回答。对话1更自然。}\\ %第七行
        \hline
         \multicolumn{2}{l}{\textbf{Translation of Assessments}} \\
        \hline
        \multicolumn{2}{p{13cm}}{\underline{Professional Counselor A}: Dialogue 1 feels more like real counseling, with emotional flow and non-verbal communication. Dialogue 2 seems more robotic, constantly answering the seeker's questions, with weaker emotional interaction and insufficient attention to the seeker's emotions.}\\ %First line
\multicolumn{2}{p{13cm}}{\underline{Professional Counselor B}: Dialogue 1 respects the seeker's decisions, offering suggestions for the seeker to make their own decisions; Dialogue 2 still focuses more on problem-solving. Counselors should not make decisions for the seeker.}\\ %Second line
\multicolumn{2}{p{13cm}}{\underline{Psychology Student C}: Dialogue 1 is better, with empathy and attempts to get the seeker to suggest effective methods they have tried before.}\\ %Third line
\multicolumn{2}{p{13cm}}{\underline{Psychology Student D}: Dialogue 2 is better, as it proposes specific solutions; the approach in Dialogue 1 may not be entirely appropriate.}\\ %Fourth line
\multicolumn{2}{p{13cm}}{\underline{Psychology Student E}: Dialogue 1 gives a sense of forcing the seeker to make a decision, which I find uncomfortable. Therefore, even if the solutions proposed in Dialogue 1 are potentially effective, they may not actually work and could cause secondary harm. Compared to the other Dialogue 2, this Dialogue 2 seems to lack effectiveness in comfort, also saying some "correct but useless" words.}\\ %Fifth line
\multicolumn{2}{p{13cm}}{\underline{Psychology Student F}: Dialogue 1 is better, providing privacy protection on sensitive topics.}\\ %Sixth line
\multicolumn{2}{p{13cm}}{\underline{Psychology Student G}: Dialogue 1 is better, as it clearly inquires about the situation, offers specific advice, and soothes the seeker's emotions. Dialogue 2 feels more like a robotic response. Dialogue 1 is more natural.}\\ %Seventh line
\hline
        \end{tabular}
    }
    
    \caption{The expert assessments of Table~\ref{tab:Psy-insight-vs-smile}. All assessments of the case comparisons are open-source on our GitHub website.}
    \label{tab:assessment1}
    \end{CJK*}
    \end{table}

\begin{table}[H]
    \begin{CJK*}{UTF8}{gbsn}
    \centering
    \small
    \resizebox{\textwidth}{!}{
        \begin{tabular}{p{6.5cm}p{6.5cm}}
        \hline
        \multicolumn{2}{l}{\textbf{Chinese Counseling Case}}\\
         \hline
        \textbf{Psy-Insight} & \textbf{Smile\cite{qiu2023smile}} \\
        \hline
        \multicolumn{2}{l}{\textbf{Source}}\\
        
        \hline
         Textbook\& Psychological blogs &  ChatGPT-Synthesis from PsyQa  \\
         \hline
         \multicolumn{2}{l}{\textbf{Topic}}\\
        \hline
        \multicolumn{2}{l}{Depression}\\
        \hline
\underline{来访者}：说实话，因为我自己无法解决自己的问题，所以我需要得到专业的帮助，这一点让我觉得很沮丧。 & \underline{来访者}：我最近总感觉心情很低落，没有办法集中精力。\\ %第一行
\underline{治疗师}：如果你寻求外界的帮助却不觉得沮丧的话，那我才要吃惊了。这是一种健康的感受。但是事实上，你已经迈出了重要的一步，这是值得庆贺的进步。这一定需要极大的勇气。 & \underline{治疗师}：听起来你可能是焦虑或者抑郁了。你有考虑过找专业人士帮助你么？\\% 第二行
\underline{来访者}：嗯，我从来没有想过这就算是一个大进步。 & \underline{来访者}：我其实很害怕去看心理医生，总觉得那是疯子才去的地方。\\ %第三行
\underline{治疗师}：现在你坐在这里，你是否相信治疗能够给你带来帮助呢？你是否相信治疗能在很短的时间内改善你的生活呢？ & \underline{治疗师}：真正勇敢的人会主动接受并寻求帮助，这句话也适用于心理健康。这并不是什么可耻的事情。\\ %第四行
\underline{来访者}：我来到这里就是因为我觉得这能帮助我。 & \underline{来访者}：可是我真的不知道该怎么办，感觉自己好像陷入了一个困境。\\ %第五行
\underline{治疗师}：我了解了你的过去，我们也已经进行了一段时间的交谈，我相信你会感觉好些的。我不会做出什么保证，但是我想，改善你的生活并非不可能。我这样想并非基于一个无意义的愿望，而是基于你给出的四个关键信息：你已经迈出了重要一步一你来到了这里，你在一定程度上相信你的生活会得到改善。此外，你已经为自己建立了一个目标：得到新的生活。最后，你对自己的痛苦能保持一种开放性的态度。换句话说，你可以描述出你的痛苦来。 & \underline{治疗师}：放心，我们一起想办法。你可以尝试一些运动或者是学习新的知识让自己分散注意力。同时我们也可以找专业人士寻求帮助。\\ %第六行
\underline{来访者}：那意思就是我还是有希望的。 & \underline{来访者}：你说的有道理，我会试着去做的。可是我总是觉得如果我失败了，再次陷入低谷，那我就真的一无所有了。\\ %第七行
\underline{治疗师}：是的，的确有希望。我还相信你很快就能感觉好起来，只需要一个条件。 & \underline{治疗师}：不要害怕失败，每个人都有过不如意的经历。但是你可以从失败中寻找经验，继续努力前行。并且不要忘记，你永远不是孤身一人，你身边有很多支持你的人。\\ %第八行
\underline{来访者}：什么条件？ & \underline{来访者}：谢谢你，现在觉得好多了。我会尽快去做的。\\ %第九行
\underline{治疗师}：那就是你愿意付出势力，哪怕面对十分困难的工作时也会势力不止。这项困难的工作听起来会很奇怪，那就是你要尽力不再那么费力去和痛苦斗争。 & \underline{治疗师}：没关系的，把握好自己的节奏，慢慢来就好。需要的时候，我一直在你的身边。\\ %第十行
\underline{来访者}：我愿意尝试一下。 &  \\ %第十一行
% \vspace{0.2cm}
 \hline
        
        \end{tabular}
    }
    
    \caption{The Chinese case comparison of Psy-insight and Simle dataset on the similar topic: Distress. Table \ref{tab:Translation2} shows its English translation.}
    % \label{tab:Psy-insight-vs-smile}
    \end{CJK*}
    \end{table}

\begin{table}[H]
    \begin{CJK*}{UTF8}{gbsn}
    \centering
    \small
    \resizebox{\textwidth}{!}{
        \begin{tabular}{p{6.5cm}p{6.5cm}}
         \hline
        \textbf{Psy-Insight} & \textbf{Smile\cite{qiu2023smile}} \\
                 \hline
        \multicolumn{2}{l}{\textbf{Dialog ID}} \\
        \hline
        \textbf{1} & \textbf{2} \\
        \hline
         \multicolumn{2}{l}{\textbf{Experts’ Assessments}} \\
        \hline
        \multicolumn{2}{p{13cm}}{\underline{专业咨询师A}：感觉1更能调动求助者的求助动机。另外，咨询师只能评估来访者的状态，不会直接说你可能焦虑或抑郁了。但2后面说“你永远不是孤身一人”“需要的时候，我一直在你的身边”比较温暖。}\\ %第一行
\multicolumn{2}{p{13cm}}{\underline{专业咨询师B}：对话1更好，对话2首次对话中仅由“心情低落，无法集中精力”就回应求助者“可能是焦虑或者抑郁”，有点过于草率，以及“真正勇敢的人会主动接受或寻求帮助”这句话有点把不敢正视心理问题归因于“不够勇敢”，不如对话1中承认一些负面情绪是自然的这样的表述好。}\\ %第二行
\multicolumn{2}{p{13cm}}{\underline{心理系学生C}：对话1更好，对话2更加自然，因为对话一探讨问题的角度更加深入，更像心理咨询，而对话2更像是日常放松状态下人们的对话。}\\ %第三行
\multicolumn{2}{p{13cm}}{\underline{心理系学生D}：1更好，1有共情，并给出了具体建议，不过1的建议有些模式化口号化，感觉有效性有限。2虽然也有建议，但是ai味较重，感觉不是特别真诚。1 的对话比较自然。但是对于一个求助者来说，1可能也不足够亲近。}\\ %第四行
\multicolumn{2}{p{13cm}}{\underline{心理系学生E}：1更好。虽然两个都有一点说教的生硬，但是1的安慰相对更多，注意肯定了来访的行为，而不是像2的祈使句命令提建议。}\\ %第五行
\multicolumn{2}{p{13cm}}{\underline{心理系学生F}：2更好，1太夸张了。}\\ %第六行
\multicolumn{2}{p{13cm}}{\underline{心理系学生G}：对话1更好，支持者对求助者的话会给出总结，并且很明显地在引导对话，对话2里说求助者可能是焦虑或抑郁有些过于直接和冰冷了。对话1更自然。}\\ %第七行
        \hline
                 \multicolumn{2}{l}{\textbf{Translation of Assessments}} \\
        \hline
        \multicolumn{2}{p{13cm}}{\underline{Professional Counselor A}: Dialogue 1 feels more capable of motivating the seeker's help-seeking motivation. Additionally, counselors can only assess the state of the visitor and will not directly say that you may be anxious or depressed. However, Dialogue 2 later says "You are never alone" and "I am always by your side when you need me," which feels warmer.}\\ %First line
\multicolumn{2}{p{13cm}}{\underline{Professional Counselor B}: Dialogue 1 is better. In the first dialogue of Dialogue 2, the response to the seeker's statement "feeling low and unable to concentrate" is "you may be anxious or depressed," which seems a bit hasty. Also, the phrase "truly brave people will actively accept or seek help" somewhat attributes the avoidance of mental issues to "not being brave enough." This is not as good as the acknowledgment in Dialogue 1 that some negative emotions are natural.}\\ %Second line
\multicolumn{2}{p{13cm}}{\underline{Psychology Student C}: Dialogue 1 is better. Dialogue 2 feels more natural because Dialogue 1 explores the problem more deeply, resembling psychological counseling, while Dialogue 2 feels more like casual conversations in a relaxed state.}\\ %Third line
\multicolumn{2}{p{13cm}}{\underline{Psychology Student D}: Dialogue 1 is better. It shows empathy and provides specific advice, although some of the advice in Dialogue 1 is somewhat formulaic and slogan-like, feeling limited in effectiveness. Dialogue 2, while also offering advice, has a stronger AI flavor, feeling less sincere. Dialogue 1's conversation is more natural. However, for a seeker, Dialogue 1 may not be close enough.}\\ %Fourth line
\multicolumn{2}{p{13cm}}{\underline{Psychology Student E}: Dialogue 1 is better. Although both have a bit of didactic stiffness, Dialogue 1 offers more comfort, paying attention to and affirming the visitor's behavior, unlike Dialogue 2's imperative sentences commanding suggestions.}\\ %Fifth line
\multicolumn{2}{p{13cm}}{\underline{Psychology Student F}: Dialogue 2 is better. Dialogue 1 is too exaggerated.}\\ %Sixth line
\multicolumn{2}{p{13cm}}{\underline{Psychology Student G}: Dialogue 1 is better. The supporter in Dialogue 1 provides summaries of the seeker's words and clearly guides the conversation. Saying in Dialogue 2 that the seeker may be anxious or depressed is too direct and cold. Dialogue 1 is more natural.}\\ %Seventh line
\hline
        \end{tabular}
    }
    \vspace{-0.2cm}
    \caption{The expert assessments of human counseling and AI-synthesized counseling. AI-generated counseling dialogues often lack empathy, interaction, and focus too much on problem-solving.}
    \label{tab:Comparison_simle_vs_psy_insight}
    \end{CJK*}
    \end{table}
\section{Translation of Previous Cases}

\begin{table}[H]
    \centering
    \small
    \resizebox{\textwidth}{!}{
        \begin{tabular}{p{8cm}p{1.7cm}p{3cm}}
         \hline
        \textbf{Chat Stage} & \textbf{Case ID} & \textbf{Psychotherapy} \\
        \hline
        The 4th Session  & 23  &  Postmodern Therapy\\
        \hline
        \textbf{Background} &  \multicolumn{2}{l}{\textbf{Guide}} \\
        \hline
        The visitor is a young woman facing challenges related to family and intimate relationships, seeking support and understanding in emotional cognition and complexity. She may be seeking solutions to dilemmas in her family and marriage, hoping for clearer insight into her emotions and choices. & \multicolumn{2}{p{5.2cm}}{The therapist needs to help her separate issues through gradual questioning, alleviate negative emotions, and clarify the impact of these issues to enhance her engagement and coping abilities in life.} \\
        \hline
        \multirow{2}*{\textbf{Dialogue In Psy-Insight}} & \textbf{Strategy 
         /} & \textbf{Reasoning /} \\
         & \textbf{Emo-label} & \textbf{Observation} \\
        \hline
            \textbf{Therapist}: You're willing to challenge your fears and be more open in expressing yourself, but you still prefer to take it step by step, right? & Question & The therapist guides the visitor to clarify issues through gradual questioning. \\
            \textbf{Client}: Of course. & Neutral & None\\
            \textbf{Therapist}: Where did you learn how to be a woman, wife, and mother? & Question & Exploring the visitor's gender role perceptions through questioning.\\
            \textbf{Client}: I don't know. I've never thought about that question. & Neutral & The visitor lacks reflection on this topic.\\
            \textbf{Therapist}: Well, I'm curious where you learned to care for others, putting your own needs behind theirs, sometimes even sacrificing your own self to fulfill others? & Question & Further probing into the origins of the visitor's self-sacrificial and others-first concepts.\\
            \textbf{Client}: (Still thinking) I guess I learned it from my parents, especially from how my father treated my mother. & Others & The visitor recognizes that her behavioral patterns and beliefs may stem from parental influence.\\
            \textbf{Therapist}: Don't you think your current way of life is learning from your mother—about how to be a woman, wife, and mother? & Question & Prompting the visitor to reflect on whether she is replicating her mother's life patterns.\\
            \textbf{Client}: I'm not sure. I've never thought about whether I'm repeating my mother's way of life. I know she never thought about going out to work or studying. From this perspective, there are significant differences between us. & Neutral & The visitor expresses uncertainty and points out significant differences between herself and her mother.\\
            \textbf{Therapist}: Yes, that does seem like a real difference. I wonder if you can find any similarities between you and her? & Question & Acknowledging the differences, the therapist encourages the visitor to identify similarities with her mother's way of life.\\
            \textbf{Client}: Well, I think my mother is very traditional. In her view, men are the leaders of the family, the primary breadwinners, and if you want, they can also be the protectors of the family. The role of women is to raise children, take care of the entire family, and I guess, including taking care of their husbands. & Neutral & The visitor describes her mother's traditional views and gender role division.\\
            \textbf{Therapist}: Among the many beliefs your mother taught you, how many have you accepted? & Question & The therapist asks about the degree to which the visitor has internalized her mother's beliefs.\\

        \hline
         \multicolumn{3}{l}{\textbf{Summary}} \\
        \hline
          \multicolumn{3}{p{11.2cm}}{As a counselor, I should guide the visitor to question, reflect, and gradually replace these unhealthy cultural beliefs, establishing a more positive self-awareness.} \\
        \hline
        \end{tabular}
    }
    \caption{The English translation of Table ~\ref{tab:chinese_example}.}
    \label{tab:translation}
    \end{table}

\begin{table}[H]
    \begin{CJK*}{UTF8}{gbsn}
    \centering
    \small
    \resizebox{\textwidth}{!}{
        \begin{tabular}{p{6.5cm}p{6.5cm}}
        \hline
        \multicolumn{2}{l}{\textbf{Chinese Counseling Case}}\\
         \hline
        \textbf{Psy-Insight} & \textbf{Smile\cite{qiu2023smile}} \\
        \hline
        \multicolumn{2}{l}{\textbf{Source}}\\
        
        \hline
         Textbook\& Psychological blogs &  ChatGPT-Synthesis from PsyQa  \\
         \hline
         \multicolumn{2}{l}{\textbf{Topic}}\\
        \hline
        \multicolumn{2}{l}{Depression}\\
        \hline
\underline{Client}: To be honest, I feel frustrated because I can't solve my own problems, so I need professional help. & \underline{Client}: I've been feeling very low lately and unable to concentrate.\\ %First line
\underline{Counselor}: If you didn't feel frustrated seeking outside help, I would be surprised. This is a healthy feeling. But in fact, you've taken an important step, which is a commendable progress. It must have taken great courage. & \underline{Counselor}: It sounds like you might be anxious or depressed. Have you considered seeking help from a professional?\\%Second line
\underline{Client}: Well, I never thought of that as a big step. & \underline{Client}: I'm actually afraid of seeing a psychologist, always felt it's a place for crazy people.\\ %Third line
\underline{Counselor}: Now that you're here, do you believe that counseling can help you? Do you believe that counseling can improve your life in a short time? & \underline{Counselor}: Truly brave people will actively accept and seek help, this applies to mental health as well. It's nothing to be ashamed of.\\ %Fourth line
\underline{Client}: I came here because I believe it can help me. & \underline{Client}: But I really don't know what to do, I feel like I'm stuck in a rut.\\ %Fifth line
\underline{Counselor}: I understand your past, and we've had a period of conversation, I believe you'll feel better. I won't make any promises, but I think improving your life is not impossible. I think this way not based on a meaningless wish, but based on the four key information you've given: you've taken an important step - you're here, you somewhat believe your life will improve. In addition, you've set a goal for yourself: to get a new life. Finally, you can maintain an open attitude towards your pain. In other words, you can describe your pain. & \underline{Counselor}: Don't worry, we'll figure out a way together. You can try some exercise or learn new knowledge to distract yourself. We can also seek help from professionals.\\ %Sixth line
\underline{Client}: So there's still hope for me. & \underline{Client}: You're right, I'll try to do that. But I always feel that if I fail and fall into a low again, I'll truly have nothing left.\\ %Seventh line
\underline{Counselor}: Yes, there is indeed hope. I also believe you'll feel better soon, with just one condition. & \underline{Counselor}: Don't be afraid of failure, everyone has had unsatisfactory experiences. But you can find lessons from failure and continue to strive forward. And don't forget, you're never alone, there are many people around you who support you.\\ %Eighth line
\underline{Client}: What condition? & \underline{Client}: Thank you, I feel much better now. I'll do it as soon as possible.\\ %Ninth line
\underline{Counselor}: That is, you're willing to exert effort, even when facing very difficult tasks. This difficult task sounds strange, that is, you should try not to struggle so hard with your pain. & \underline{Counselor}: It's okay, take your own pace, take it slow. I'm always here when you need me.\\ %Tenth line
\underline{Client}: I'm willing to try. &  \\ %Eleventh line
% \vspace{0.2cm}
 \hline
        
        \end{tabular}
    }
    
    \caption{The English translation of Table ~\ref{tab:Comparison_simle_vs_psy_insight}.}
    \label{tab:Translation2}
    \end{CJK*}
    \end{table}

\section{Annotation Guideline for Volunteers}
\label{sec:Guideline}

\begin{table}[H]
    \centering
    
    \label{tab:human-eval-guideline}
    
    \begin{tabular}{p{\textwidth}}
    \toprule
    \textbf{Guideline For human evaluation} \\
    \hline
    This study aims to evaluate text generation systems with various mental health counseling questions as system prompts. Each entry contains 4 parts: question title, description, label, and answer text. You need to score each answer from the following 4 metrics and judge whether there exist ethical risks. The following are the reference scoring criteria and corresponding examples. \\
    \hline
    \textbf{Interaction} -- As the patient, are you willing to continue the consultation? ... \\
    \textbf{e.g.}\\
    (Low interaction) The response is long and wordy or general. \\
    (High interaction) The AI asks questions that facilitate the conversation/gives targeted suggestions.\\ 
    \hline
    \textbf{Helpfulness} -- If you were the patient, would this response solve your problem? \\
    \textbf{e.g.} \\
    (High helpfulness) Responses to patients’ questions that provide targeted suggestions.\\ 
    \hline
    \textbf{Comforts} -- If you put yourself in the patient's shoes, would this response make you feel comforted? \\
    \textbf{e.g.} \\
    (High comfort) The reply expressed sympathy \\
    (Low comfort) The reply asked a stupid question\\ 
    \hline
    \textbf{Explainability} -- Put yourself in the shoes of a consultant and pay attention to judging from the consultant's perspective whether the model's explanation of the response is understandable? Can you understand the intention behind the model's generated response? \\
    \textbf{e.g.} \\
    (Interpretability) Combined with the explanation information generated by the partial model and the conversation context, can you understand how this response contributes to the consultation conversation?\\ 
    \hline
    \end{tabular}
\caption{Evaluation guideline for rater.}
\label{tab:guideline}
\end{table}

\end{document}